%% file: main.tex
\pdfoutput=1

\documentclass[11pt]{article}

\usepackage{ACL2023}

\usepackage{times}
\usepackage{latexsym}

\usepackage[T1]{fontenc}

\usepackage[utf8]{inputenc}

\usepackage{microtype}
\usepackage{inconsolata}

\usepackage{inconsolata}
\usepackage{microtype}
\usepackage{rotating}
\usepackage{blindtext}
\usepackage{amsmath}
\usepackage{amsfonts}
\usepackage{graphicx}
\usepackage{mathtools}
\usepackage{booktabs}
\usepackage{relsize}
\usepackage{float}
\usepackage{graphics}
\usepackage{multirow}
\usepackage{algorithm}
\usepackage{algpseudocode}
\usepackage[utf8]{inputenc}
\usepackage{tabularx}

\DeclareUnicodeCharacter{01B0}{\`u}
\DeclareUnicodeCharacter{01A1}{\`o}

\usepackage{threeparttable}
\usepackage{booktabs, caption, makecell}

\usepackage[title]{appendix}
\usepackage[cal=cm]{mathalfa}
\usepackage[colorinlistoftodos]{todonotes}
\usepackage{multirow}
\usepackage[export]{adjustbox}
\usepackage{subcaption}
\usepackage{pifont}
\newcommand{\metric}{RQUGE}
\newcommand{\mycomment}[1]{}                     

\title{\metric: Reference-Free Metric for Evaluating Question Generation by Answering the Question}

\author{Alireza Mohammadshahi\thanks{~~Work done during an internship at Meta AI.}~~$^{1,2,3}$ ~~Thomas Scialom$^1$ ~~Majid Yazdani\thanks{~~This work is done when the author was on leave from BYJU's LAB.}~~$^4$ \\ \textbf{Pouya Yanki}$^1$ ~~\textbf{Angela Fan}$^1$ ~~\textbf{James Henderson}$^2$ ~~ \textbf{Marzieh Saeidi}$^1$ \vspace{0.1cm}\\ 
 $^1$ Meta AI ~~~~~ $^2$ IDIAP Research Institute ~~~~~ $^3$ EPFL ~~~~~ $^4$ BYJU's LAB \\
 \texttt{\{alireza.mohammadshahi,james.henderson\}@idiap.ch}\\\texttt{\{tscialom,pya,angelafan,marzieh\}@meta.com} \\\texttt{\{majid.yazdani\}@byjus.com} 
}

\begin{document}
\maketitle

\input{abstract.tex}

\input{introduction.tex}
\input{relatedwork.tex}
\input{model.tex}

\input{implementation.tex}
\input{results.tex}
\input{conclusion.tex}
\input{ack.tex}
\input{limitations.tex}

\bibliography{custom}
\bibliographystyle{acl_natbib}

\appendix
\input{appendix.tex}

\end{document}

%% file: abstract.tex
\begin{abstract}
    Existing metrics for evaluating the quality of automatically generated questions such as BLEU, ROUGE, BERTScore, and BLEURT compare the reference and predicted questions, providing a high score when there is a considerable lexical overlap or semantic similarity between the candidate and the reference questions. This approach has two major shortcomings. First, we need expensive human-provided reference questions. Second, it penalises valid questions that may not have high lexical or semantic similarity to the reference questions. In this paper, we propose a new metric, \metric, based on the answerability of the candidate question given the context. The metric consists of a question-answering and a span scorer modules, using pre-trained models from existing literature, thus it can be used without any further training. We demonstrate that \metric\ has a higher correlation with human judgment without relying on the reference question. Additionally, \metric\ is shown to be more robust to several adversarial corruptions. Furthermore, we illustrate that we can significantly improve the performance of QA models on out-of-domain datasets by fine-tuning on synthetic data generated by a question generation model and re-ranked by \metric.\footnote{The implementation and annotated data are provided at \url{https://github.com/alirezamshi/RQUGE}}
\end{abstract}
\begin{figure}[!htb]
    \centering
    \includegraphics[width=\linewidth]{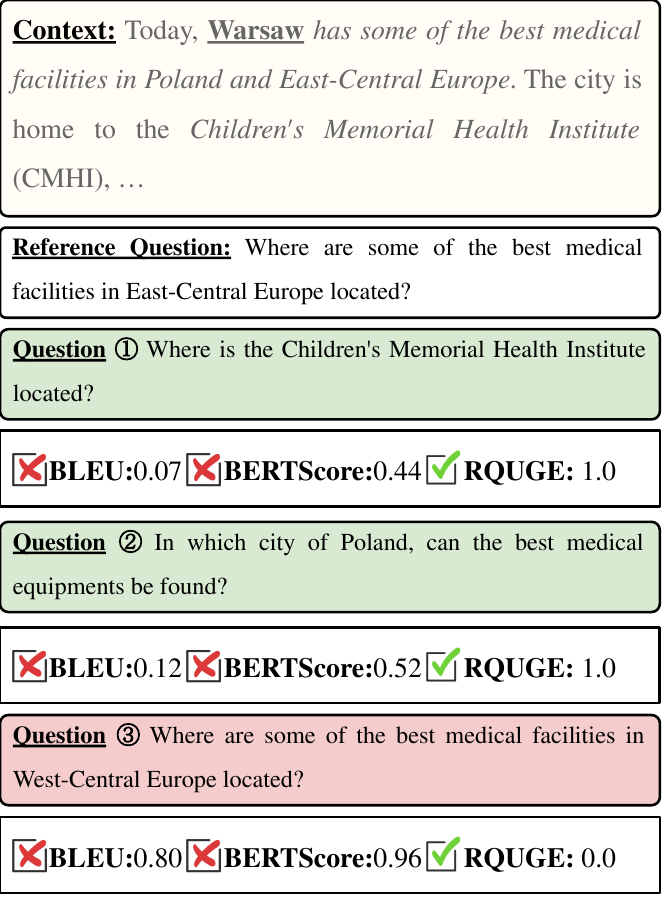}
    \caption{Normalised scores for different candidate questions. Metrics based on similarity to a reference question can penalise valid candidate questions, and compute a high score for unacceptable questions that are lexically similar to the reference. This can lead to the failure of reference-based metrics for valid questions, such as $Q_1$. Additionally, even paraphrases of the reference, like $Q_2$, may receive low scores. Furthermore, reference-based metrics may not detect small corruptions or variations in the reference, such as $Q_3$.
    }
    \label{fig:motivation}
\end{figure}

%% file: introduction.tex
\section{Introduction}

Given the context~(e.g. paragraph), the goal of question generation~(QG) is to generate questions with or without providing the answer spans. Automatic question generation can be used in several applications: improving the question answering~(QA) task~\cite{duan-etal-2017-question,du-cardie-2018-harvesting,puri-etal-2020-training,cheng-etal-2021-guiding}, automatic assessments~\cite{rebuffel-etal-2021-data, lee2021qace}, especially for the educational domain~\cite{Chen_Yang_Hauff_Houben_2018}, and the evaluation of factual consistency in the text generation tasks~\cite{scialom2019answers,scialom2021questeval,fabbri-etal-2022-qafacteval}. 

Previous work~\cite{hosking-riedel-2019-evaluating,scialom2019self, zhang-bansal-2019-addressing, laban-etal-2022-quiz} has
shown that QG models can generate questions inconsistent with the corresponding context and the answer span. So, measuring the acceptability of candidate questions is a critical challenge. Human judgment is the most accurate method in natural language generation, but it is expensive, time-consuming, and not scalable. Consequently, several metrics e.g. BLEU~\cite{papineni-etal-2002-bleu}, ROUGE~\cite{lin-2004-rouge}, BERTScore~\cite{Zhang2020BERTScore} are proposed to automatically measure the quality of the generated text. 

Specifically for the question generation task, previous work has utilised reference-based metrics e.g. BLEU, ROUGE, BERTScore, and BLEURT~\cite{sellam-etal-2020-bleurt,ushio-etal-2023-an-empirical,ushio-etal-2023-a-practical-toolkit,ushio2022generative} to evaluate the quality of the candidate question given the reference question. However, these methods highly depend on the diversity of the reference questions for a given answer span. Due to the huge cost of human annotations, existing QA/QG datasets mostly provide one reference question for the given context and answer, which results in wrongly penalising some valid questions. In Figure~\ref{fig:motivation}, the first candidate question~($Q_1$) is generated by paying attention to different evidence in the context, and $Q_2$ is a paraphrase of the reference, but both BLEU and BERTScore fail to assign high scores to them. Furthermore, reference-based metrics are not sensitive to very small corruptions of the reference questions, which makes the candidate question unacceptable~($Q_3$).

\begin{figure*}
  \centering
  \includegraphics[width=\linewidth]{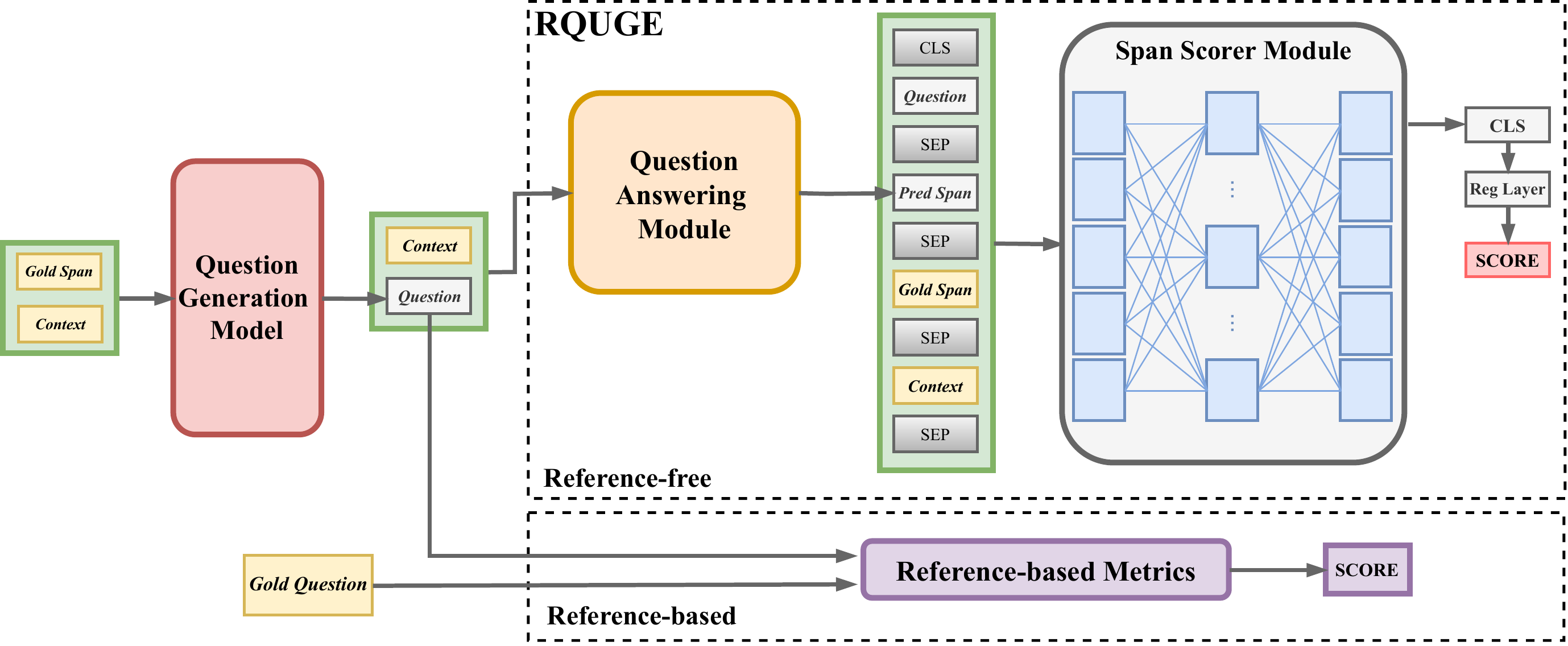}
  \caption{The architecture of \metric\ metric~(upper-side) for the question generation task, which consists of a question answering and a span scorer modules to compute the acceptability of the candidate question. Reference-based metrics are also shown at bottom of the figure, where the score is calculated by comparing the gold and predicted questions.}
  \label{fig:rquge_model}
\end{figure*}

In this paper, we propose \metric\ , a \underline{\textbf{R}}eference-free \underline{\textbf{QU}}estion \underline{\textbf{G}}eneration \underline{\textbf{E}}valuation metric that can compute the quality of the candidate question without requiring a reference question. Given the corresponding context and answer span, our metric calculates the acceptability score by applying a general question-answering module, followed by a span scorer. The former module generates the answer span for the given candidate question, and the latter computes the semantic similarity of the predicted and gold answer spans. Our metric is extremely valuable in cases where the reference question is not well-formed~\footnote{For instance, many questions in the Natural Question dataset~\cite{kwiatkowski-etal-2019-natural} are not well-formed. Example: ``in a deep mine temperatures increase with depth at the rate of''} or there is one~(or no) reference for a given context and answer span. \\
We evaluate our metric on several datasets, including SQuAD~(v1)~\cite{rajpurkar-etal-2016-squad}, Natural Questions (NQ)~\cite{kwiatkowski-etal-2019-natural}, and MS-MARCO~\cite{msmarco2021}, and show that it consistently has a better correlation with human judgment compared to previous QG metrics. We also integrate \metric\ into the decoding step by re-ranking candidate questions of each instance by our metric, leading to a better correlation with the human evaluation. Additionally, we demonstrate that \metric\ is more robust to adversaries than previous metrics with +13.1\% relative improvement. Finally, we improve the performance of question answering models on an out-of-domain dataset by fine-tuning them on synthetic data generated by a question generation model, then re-ranked with \metric\ to choose the best candidate question for the given answer span, resulting in an +18.3\% F1 and +22.2\% EM relative improvement. \\
To sum up, our contributions are as follows:
\begin{itemize}
    \item We propose \metric , an evaluation metric for measuring the quality of the automatically generated questions, without requiring access to any reference questions.
    \item We show that our metric has a significantly higher correlation with human judgment in terms of the acceptability of the candidate questions on SQuAD~(v1), NQ, and MS-MARCO datasets. Also, re-ranking candidate questions with \metric\ leads to a better correlation with human judgment.
    \item We demonstrate that \metric\ metric is more robust compared to previous work on several adversarial strategies such as negation, entity swapping, gender reversing, or paraphrasing the reference questions.
    \item Finally, we illustrate that the performance of QA models significantly improves on the out-of-domain datasets by fine-tuning them on the synthetic data, created by applying a question generator model, then re-ranking with \metric\ metric. 
\end{itemize}

%% file: relatedwork.tex
\section{Related Work}

Previous work on automatic evaluation of Natural Language Generation (NLG) tasks have been categorized as follows:

\paragraph{Unsupervised Metrics.} It contains the most commonly used metrics e.g. BLEU~\cite{papineni-etal-2002-bleu}, ROUGE~\cite{lin-2004-rouge}, chrF~\cite{popovic-2015-chrf}, and METEOR~\cite{denkowski-lavie-2010-extending}. These metrics calculate the correlation of reference and predicted sequences in a discrete space by utilising token-level matching functions. Then, recent work e.g. BERTScore~\cite{Zhang2020BERTScore} and MoverScore~\cite{zhao-etal-2019-moverscore} use BERT~\cite{devlin-etal-2019-bert} embeddings to provide a soft token-level matching instead of the hard n-gram overlap. These metrics have been applied to various NLG tasks~\cite{du-etal-2017-learning,https://doi.org/10.48550/arxiv.1704.01792,xiong-etal-2019-tweetqa,pan-etal-2020-semantic,lewis-etal-2020-bart,cheng-etal-2021-guiding,mohammadshahi2022small100,mohammadshahi2022compressed}. Specifically for QG evaluation, \citet{nema-khapra-2018-towards} propose a scoring function, focusing on the \textit{answerability} of the candidate question, which improves the human correlation when integrated with existing unsupervised metrics.

\paragraph{Regression-based Metrics.} These metrics e.g. COMET~\cite{rei-etal-2020-comet}, BLEURT~\cite{sellam-etal-2020-bleurt}, S3~\cite{peyrard-etal-2017-learning}, and VRM~\cite{HIRAO20071521} train a regression layer in a supervised manner to mimic the human judgment.

\paragraph{Ranking-based Metrics.} The aim of these metrics is to assign a higher score to a better candidate compared to worse predictions. The most popular ones include BEER~\cite{stanojevic-simaan-2014-beer} and COMET~\cite{rei-etal-2020-comet}. 

\paragraph{Generation-based Metrics.} The idea is to formulate the evaluation of NLG, as a text generation problem from pre-trained language models. Given the source sequence, the better candidate should be generated with a higher score~(probability) compared to the worse ones. The most popular ones are BARTScore~\cite{bartscore} and PRISM~\cite{thompson-post-2020-automatic}. 

Additionally, we include CTC~\cite{deng-etal-2021-compression} and QRelScore~\cite{wang2022qrelscore} as reference-free metrics for better comparison. CTC~\cite{deng-etal-2021-compression} proposes an evaluation framework for NLG tasks by providing several reference-free metrics, which are computed by aggregating the alignment scores between the input, context and the predicted sequence. To measure the alignment, CTC~\cite{deng-etal-2021-compression} uses BERT~\cite{devlin-etal-2019-bert} embedding matching, discriminative, and regression models.~\footnote{CTC can be considered as an unsupervised metric when BERT embeddings are used to compute the alignment.} QRelScore computes the answerability of the candidate question by applying word-level hierarchical matching and sentence-level prompt-based generation. Different from previous work, \metric\ combines question answering and span scoring modules to compute the acceptability of the candidate, which leads to a significantly better correlation with human judgement in multiple datasets with different domains and answer lengths.


%% file: model.tex
\section{\metric\ Architecture}

\metric\ architecture is illustrated in Figure~\ref{fig:rquge_model}. It consists of two components: question answering and span scorer modules. Given the context, gold answer span, and the candidate question, generated by a question generation model~($\operatorname{QG}$), \metric\ computes the acceptance score~($\kappa$) of the candidate question as follows:
\begin{align}
\label{eq:rquge_main}
\begin{cases}
a_c = \operatorname{QA}(q_c,D) \\
\kappa = \operatorname{S}(q_c,a_c,a_r,D) 
\end{cases}
\end{align} 
where the $q_c = \operatorname{QG}(a_r,D)$ is the generated candidate question for the gold answer span $a_r$ and context $D$. To calculate the score, the question answering model~$\operatorname{QA(.)}$ predicts the answer span $a_c$, given the candidate question $q_c$ and the context~ $D$. Finally, the span scorer $\operatorname{S(.)}$ computes the acceptance score $\kappa$, conditioned on the candidate question, predicted answer, gold answer, and context. In the following, we will describe each module in detail.

\subsection{Question Answering Module}

Given the context and the candidate question, the question answering model predicts the answer span. To make our metric general to several domains, we use UnifiedQAv2~\cite{khashabi2022unifiedqav2} model to generate the answer span. UnifiedQAv2 is a T5-based encoder-decoder model, which is trained on 20 QA datasets, and achieves competitive performance with the state-of-the-art models in several in-domain and out-of-domain datasets.\footnote{We refer to \citet{khashabi2022unifiedqav2} for further details. A list of evaluated datasets is provided in Appendix~\ref{app:dataset_list}. We use {\tt unifiedqa-v2-t5-large} checkpoint, provided in \url{https://github.com/allenai/unifiedqa}.} The input to the model is the concatenation of the candidate question and corresponding context.

\subsection{Span Scorer Module}

Given the predicted answer span $a_c$ of the candidate question $q_c$, the span scorer calculates the score (ranging from 1 to 5) of the candidate question. Inspired by \citet{chen-etal-2020-mocha} and \citet{fabbri-etal-2022-qafacteval}, we use an encoder-only BERT-based model to calculate the acceptance score. Specifically, we first encode the input sequence, then pass the vector representation of $\operatorname{[CLS]}$ to the regression layer to compute the acceptance score $\kappa$. The input to the module is: 
\begin{align*}
\label{eq:input_span}
\begin{split}
\begin{adjustbox}{width=\linewidth}
\text{\textbf{[CLS] }} \text{cand. question } \text{\textbf{[q]}} \text{ gold answer } \text{\textbf{[r]}} \text{ pred answer } \text{\textbf{[c]}} \text{ context}
\end{adjustbox}
\end{split}
\end{align*}

\begin{table*}
\centering
\small
   \begin{adjustbox}{width=0.9\linewidth}
  \begin{tabular}{lcccccccccccc}
    \toprule
    \multirow{2}{*}{Metric} & 
      \multicolumn{3}{c}{Grammaticality} &&
      \multicolumn{3}{c}{Answerability} &&
      \multicolumn{3}{c}{Relevance}\\
      \cline{2-4} \cline{6-8} \cline{10-12}
     & $r$ & $\rho$ & $\tau$ && $r$ & $\rho$ & $\tau$  && $r$ & $\rho$ & $\tau$\\
    \midrule
    \textbf{Unsupervised} &    &  &  &\vline&   &  & &\vline&   & & \\
BLEU-4 & 0.133 & 0.096 & 0.077 &\vline& 0.273 & 0.335 & 0.258 &\vline& 0.213 & 0.235 & 0.191 \\
ROUGE-1 & 0.156 & 0.096 & 0.077 &\vline& 0.312 & 0.274 & 0.217 &\vline&  0.330 & 0.322 & 0.264 \\
ROUGE-L &0.210 & 0.148 & 0.120 &\vline&  0.321 & 0.294 & 0.233 &\vline& 0.322 & 0.316 & 0.259 \\
METEOR & 0.143 & 0.086 & 0.069 &\vline& 0.334 & 0.321 & 0.251 &\vline& 0.317 & 0.315 & 0.255 \\
QBLEU & 0.160 & 0.134 & 0.106 &\vline& 0.227 & 0.235 & 0.183 &\vline& 0.240 & 0.248 & 0.200 \\
MOVERScore &  0.161 & 0.103 & 0.082 &\vline& 0.294 & 0.318 & 0.248 &\vline& 0.280 & 0.313 & 0.254 \\
BERTScore & 0.262 & 0.203 & 0.160 &\vline& 0.336 & 0.333 & 0.260 &\vline&  0.309 & 0.311 & 0.253 \\
\midrule
\textbf{Regression-based} &    &  &  &\vline&   &  & &\vline&   & & \\
BLEURT-20 &  0.203 & 0.144 & 0.113 &\vline& 0.359 & 0.341 & 0.268 &\vline& \underline{0.363} & \underline{0.363} & \underline{0.295} \\
\midrule 
\textbf{Ranking-based} &    &  &  &\vline&   &  & &\vline&   & & \\
COMET &  \underline{0.309} & \underline{0.274} & \underline{0.215} &\vline& 0.319 & 0.312 & 0.243 &\vline&  0.300 & 0.307 & 0.248 \\
\midrule 
\textbf{Generation-based} &    &  &  &\vline&   &  & &\vline&   & & \\
BARTScore &0.212 & 0.145 & 0.115 &\vline& 0.349 & \underline{0.345} & \underline{0.269} &\vline&  0.332 & 0.323 & 0.262 \\
\midrule
\textbf{Ref-Free} &    &  &  &\vline&   &  & &\vline&   & & \\
CTC & 0.120 & 0.131 & 0.110 &\vline& 0.291 & 0.243 & 0.185 &\vline& 0.195 & 0.179 & 0.145 \\
QRelScore & 0.202 & 0.102 & 0.102 &\vline& \underline{0.366} & 0.285 & 0.22 &\vline& 0.294 & 0.212 & 0.188 \\
\midrule
\textbf{\metric\ } & \textbf{0.380} & \textbf{0.278} & \textbf{0.220} &\vline& \textbf{0.604} & \textbf{0.436} & \textbf{0.344} &\vline& \textbf{0.551} & \textbf{0.403} & \textbf{0.325} \\
    \bottomrule
  \end{tabular}
   \end{adjustbox}
  \caption{\label{tab:human} Correlation of human judgment and automatic evaluation metrics based on Pearson $r$, Spearman $\rho$, and Kendall $\tau$ correlation coefficients, averaged over subsets of SQuAD, MS-MARCO, and NQ datasets. The best and second best scores are specified with bold and underline markers.}
      \vspace{-1ex}
\end{table*}

We employ pre-trained RoBERTa model, provided by \citet{fabbri-etal-2022-qafacteval}. The model is first pre-trained with a QA-infused pre-training objective,\footnote{It includes pre-training contextual embeddings with a bi-encoder extractive QA loss, which results in encoding information about questions that can be answered by each text span. The pre-trained model is available at \url{https://github.com/salesforce/QAFactEval}. } then fine-tuned on MOCHA human ratings QA dataset~\cite{chen-etal-2020-mocha}. MOCHA is a dataset of human judgment scores for training and testing reading comprehension metrics, where annotators are asked to score candidate spans, given the context, gold answer, and the corresponding question.

%% file: implementation.tex
\section{Experimental Setup}
\label{sec:impl}

\paragraph{Datasets.} We evaluate metrics on three widely-used QA datasets, including SQuAD(v1)~\cite{rajpurkar-etal-2016-squad}, NQ~\cite{kwiatkowski-etal-2019-natural}, and MS-MARCO~\cite{msmarco2021}. NQ is used to demonstrate the benefit of our metric in cases where reference questions are not well-formed and are derived from the Google engine.~\footnote{We use the subset of NQ with \textit{short-form} answer spans, similarly to the previous work~\cite{murakhovska-etal-2022-mixqg}.} For MS-MARCO, we use {\tt DESCRIPTION} type of the dataset to show the effectiveness of our metric on candidate questions with long answer spans~(13 tokens on average). Unlike SQuAD and NQ, MS-MARCO is not included in the training data of the question answering module of \metric~(i.e. UnifiedQAv2~\cite{khashabi2022unifiedqav2}). We use MS-MARCO to demonstrate that \metric\ can be generalised to out-of-domain datasets.~\footnote{Further details of evaluated datasets are provided in Appendix~\ref{app:eval_dataset}.}

\paragraph{Question Generators.} We fine-tune two commonly used QG models, including GPT2~\cite{radford2019language}, trained with causal language modelling objective, and MixQG~\cite{murakhovska-etal-2022-mixqg}, which is the state-of-the-art question generator and is a T5-based~\cite{2020t5} sequence-to-sequence model. We choose GPT2 and MixQG as our question generators as there is a significant gap in their performance, making them suitable for evaluating the metrics.~\footnote{Hyper-parameters for fine-tuning QG models are provided in Appendix~\ref{app:hyper-finetune}.}

\paragraph{Baselines.} We include BLEU-4~\cite{papineni-etal-2002-bleu}, ROUGE-1, ROUGE-L~\cite{lin-2004-rouge}, METEOR~\cite{denkowski-lavie-2010-extending}, MoverScore~\cite{zhao-etal-2019-moverscore}, and BERTScore~\cite{Zhang2020BERTScore}\footnote{We use the better-performing BERTScore, initialised with DeBERTa-xlarge model~\cite{he2021deberta}.} as unsupervised metrics, that are commonly used for the question generation task. We additionally use QBLEU, which is specific for QG evaluation. Furthermore, we utilise BLEURT-20~\cite{sellam-etal-2020-bleurt}, and COMET~\cite{rei-etal-2020-comet} as regression-based and ranking-based metrics. Finally, we include BARTScore~\cite{bartscore}, fine-tuned on ParaBank2~\cite{hu-etal-2019-large}. In all aforementioned metrics, scores are calculated between the candidate and reference questions. As reference-free baselines, we use QRelScore~\cite{wang2022qrelscore}, and also adopt the factual consistency scorer of CTC~\cite{deng-etal-2021-compression} to calculate the consistency score as:
\begin{align*}
\kappa_{ctc} = \operatorname{mean}(\operatorname{align}([a_r,D] \rightarrow q_c))
\end{align*}
where $a_r$, $D$, and $q_c$ are answer span, context, and candidate question, respectively. The function $\operatorname{align}(.)$ estimates the alignment for tokens of the candidate question with the given context and answer. we use {\tt albert-xlarge-vitaminc-mnli}, which uses a discriminative model to compute the alignment.\footnote{We also used faithfulness aspect of BARTScore~\cite{bartscore} as a reference-free metric for measuring the quality of the candidate questions, but preliminary experiments result in a poor correlation with human judgment.} 

\paragraph{Human Annotation.} We use 3 volunteer annotators to rate the candidate questions of QG models.\footnote{Human evaluation of the quality of the generated questions is not available in previous work.} All annotators are fluent English speakers. 
Inspired by previous work~\cite{rus-etal-2010-first,nema-khapra-2018-towards}, we ask annotators to score each candidate question based on three criteria: \textit{grammaticality}, \textit{answerability}, and \textit{relevance}. Grammaticality measures the syntactic structure of the question. Answerability checks whether the question contains all the important entities, and relevance checks the relatedness of the generated questions with the given answer span. Grammaticality and answerability scores are on a 3-point scale (3 as acceptable, and 1 as rejection), and relevance is on a 2-point scale. We sample $600$ questions generated from fine-tuned QG models on SQuAD(v1)~\cite{rajpurkar-etal-2016-squad}, NQ~\cite{kwiatkowski-etal-2019-natural}, and MS-MARCO~\cite{msmarco2021} datasets. We then randomly shuffle and anonymise them for annotators. Further details of the human annotation procedure are provided in Appendix~\ref{app:human}.\footnote{The human-evaluated data is available at \url{https://github.com/alirezamshi/RQUGE}.}

%% file: results.tex
\section{Results and Discussion}
We evaluate our \metric\ and previous metrics on various datasets and tasks. First, we evaluate the correlation of metrics with human judgment in Sections \ref{result:human-eval} and \ref{result-re-rank}. We then demonstrate their robustness on the adversarial subset in Section~\ref{result:robust}. Finally, Section~\ref{result:qa-task} illustrates that fine-tuning QA models on the synthetic data, created by our metric, improves their performance on out-of-domain datasets.
\subsection{Correlation with Human Judgment}
\label{result:human-eval}

\begin{figure}
  \centering
  \includegraphics[width=\linewidth]{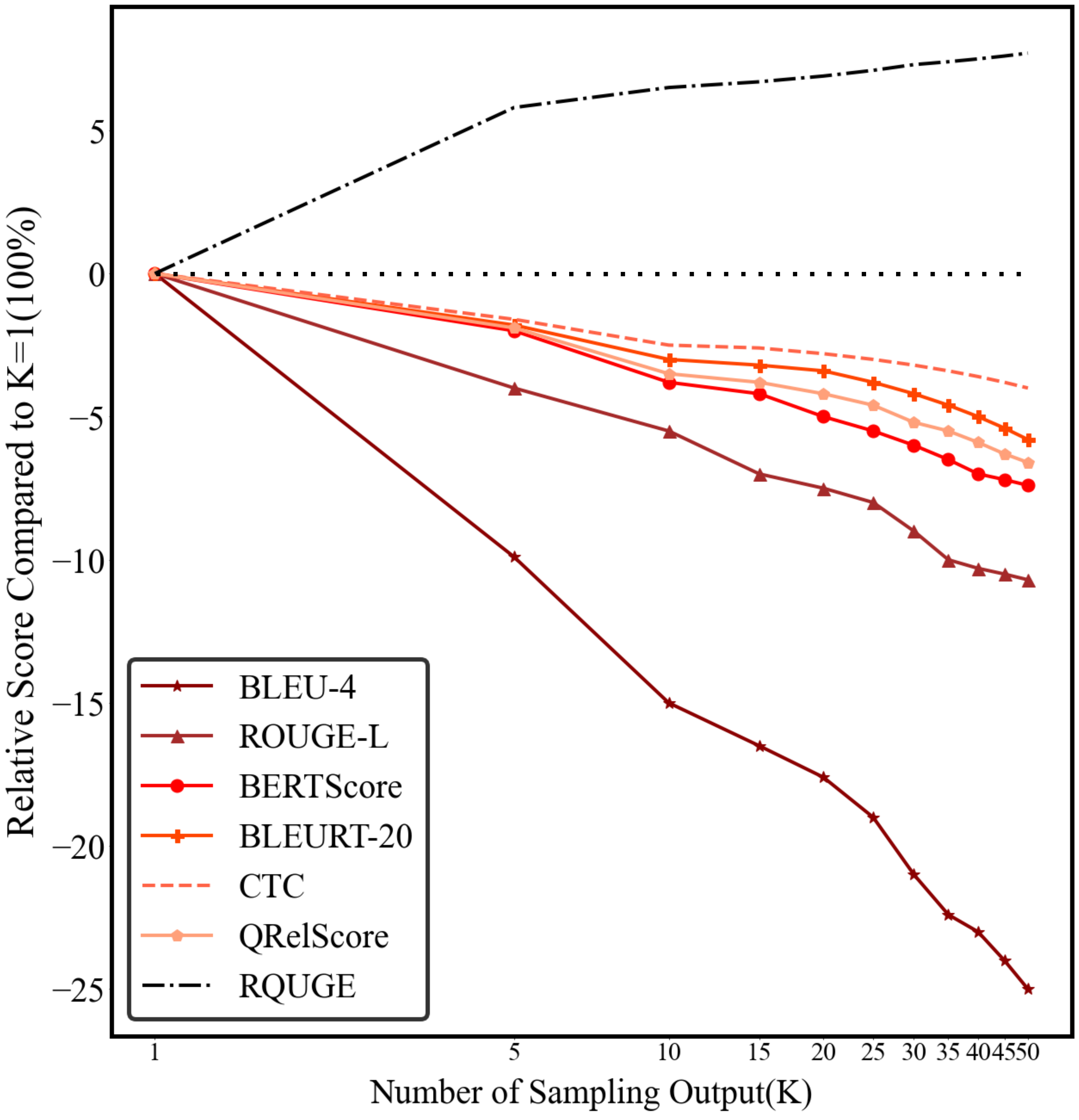}
  \caption{The relative score of automatic metrics compared to $K=1$ for different values of $K$, after re-ranking the output of the question generation model with \metric\ metric. \metric\ increases as it is the objective of the re-ranking mechanism.}
  \label{fig:squad_rerank}
\end{figure}

\paragraph{Annotator Agreement.} The pairwise inter-annotator agreements, calculated using Cohen's Kappa are 88.91\%, 85.32\%, and 83.54\%.~\footnote{Calculated on the summation of grammaticality, answerability, and relevance.} We use the average score of three annotators for the remaining experiments. 

\paragraph{Metric-to-Human Correlation.} Table~\ref{tab:human} illustrates the correlations of automatic metrics with the human judgment, averaged over all datasets.\footnote{For a fair comparison on \textit{grammaticality}, we just evaluate on SQuAD evaluation set, where reference questions are mostly well-formed.} \metric\ metric has a considerably higher correlation with human judgment on all criteria. For instance, it outperforms the best previous work with +7.1\%, +23.8\%, +18.8\% absolute improvement~(based on Pearson~($r$) score) for \textit{grammaticality}, \textit{answerability}, and \textit{relevance}, respectively. Appendix~\ref{app:result_human} illustrates the result of correlation with the human judgment for each dataset, separately. In in-domain evaluation sets, \metric\ results in +29.7\% and +24.8\% absolute point improvement for SQuAD and NQ datasets, respectively, based on answerability measurement. For MS-MARCO as the out-of-domain dataset, \metric\ reaches +12.2\% absolute improvement for the relevancy criterion, while having competitive results with CTC on answerability measurement. These results show the effectiveness of our metric in different domains, and question structures ~(well-formedness) and confirm the generalisation of our metric to out-of-domain settings.

\begin{figure}
  \centering
  \includegraphics[width=\linewidth]{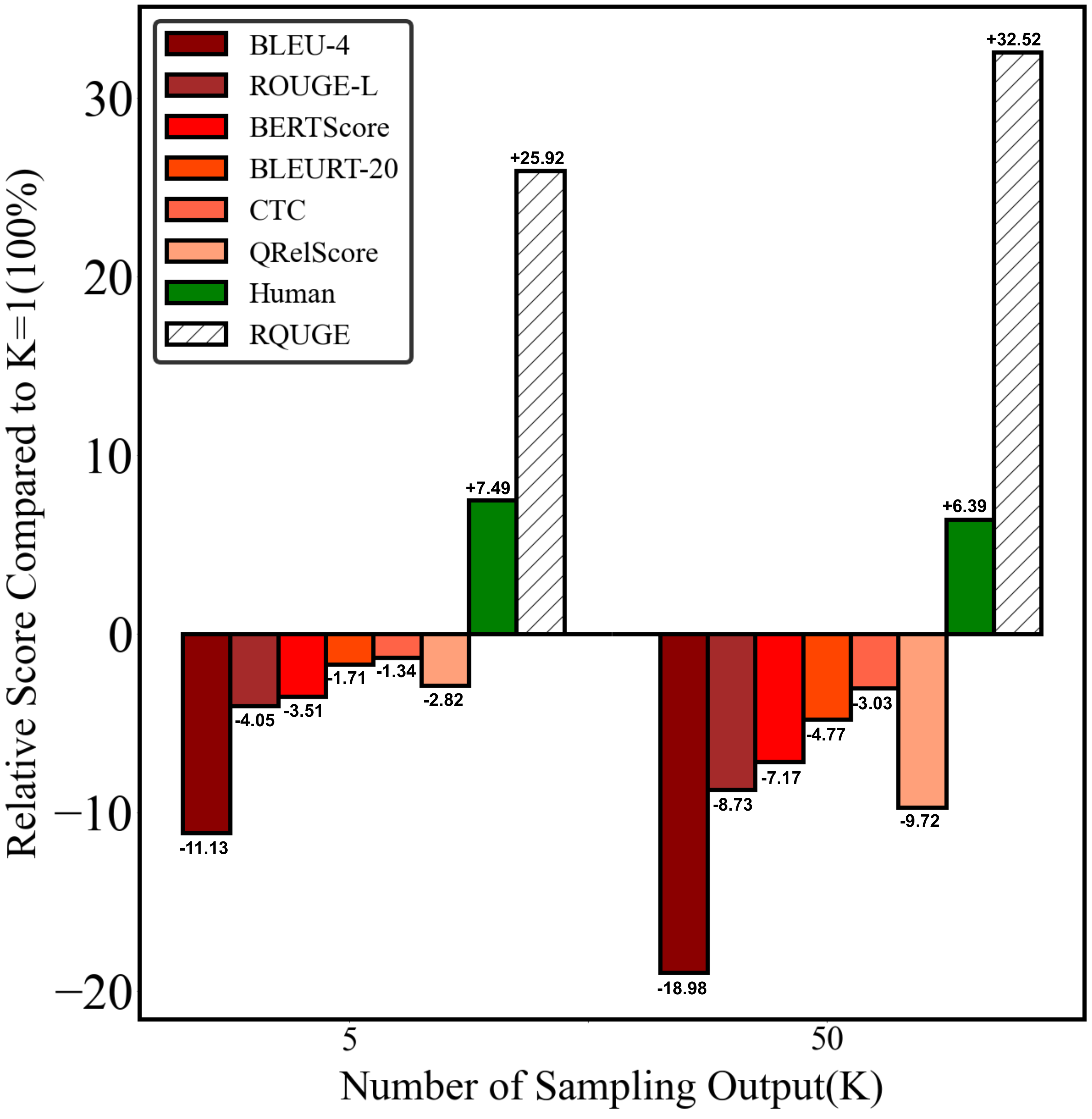}
  \caption{Relative score of automatic metrics compared to $K=1$ for 250 samples from the evaluation set of SQuAD. For the human score, we use the average score of corresponding samples. It shows that re-ranking with \metric\ gives better output~(as both $K=5,50$ have better correlation with human), while all QG metrics except \metric\ are diverging after the re-ranking.}
  \label{fig:human_rerank}
\end{figure}

\subsection{Re-Ranking with \metric}
\label{result-re-rank}
\begin{table*}
\centering
\begin{adjustbox}{width=0.8\linewidth}
\begin{tabularx}{\textwidth}{l|X|X|X}
Corruption & Context & Reference & Candidate \\
\toprule
\small{Paraphrasing} & \small{\textbf{Orange County} is a rapidly developing business center that includes Downtown Santa Ana, the South Coast Metro and Newport Center districts; ...} & \small{Which county is \textbf{developing} its business center?}  & \small{Which county is \textbf{expanding} its business center?} \\ \hline
\small{Negation} & \small{\textbf{Ondemar Dias} is accredited with first discovering the geoglyphs in 1977 and Alceu Ranzi with furthering their discovery after flying over Acre. ...} & \small{Who is given credit for discovering geoglyphs along the Amazon River?} & \small{Who is \textbf{not} given credit for discovering geoglyphs along the Amazon River?} \\ \hline 
\small{Entity Swap} & \small{..., a plague claimed some 1.7 million victims in \textbf{Italy}, ... killed about 100,000 in Sweden, and \textbf{300,000 in Prussia}. The plague killed ...} & \small{How many were killed by plague in \textbf{Italy} in the 17th century?} & \small{How many were killed by plague in \textbf{Prussia} in the 17th century?} \\ \hline
\small{Reverse Gender} & \small{... For example, \textbf{Joseph Haas} was arrested for allegedly sending an email to the Lebanon, New Hampshire city councilors stating, "Wise up or die."} & \small{What did Joseph Haas say in \textbf{his} email?} & \small{What did Joseph Haas say in \textbf{hers} email?} \\
\bottomrule
\end{tabularx}
\end{adjustbox}
\caption{\label{tab:adv-samples} Samples of the adversarial subset for evaluating the robustness of QG metrics.}
\end{table*}

To further demonstrate the effectiveness of \metric , we use it to re-rank the output predictions of the question generation model to choose the best generated question. Given the context and answer span, QG model~\footnote{We use MixQG~\cite{murakhovska-etal-2022-mixqg} model as the question generation model, as it is the state-of-the-art QG model for the evaluated dataset.} generates a bag of candidate questions~(here, we apply Nucleus sampling~\cite{Holtzman2020The} to increase the diversity)\footnote{We use temperature sampling of 1, and top summation probability~($top_p$) of 0.94.}, that are sorted based on the perplexity~(PPL) of the question generator. At each step, we choose $K$-first candidates of each sample and re-rank them with \metric. Then, other automatic metrics are computed for the best one chosen by \metric. \\
Figure~\ref{fig:squad_rerank} demonstrates the relative score gains of QG metrics compared to $K=1$~(best candidate by PPL) for different values of $K$.~\footnote{The relative performances of the remaining metrics for different $K$ are provided in Appendix~\ref{app:reranking}.} Interestingly, it illustrates that other metrics drop as the number of candidate outputs of each sample~($K$) is increasing, meaning that the best candidates chosen by our metric are not correlated with other metrics. To confirm these results, annotators are asked to score 250 samples~\footnote{We filter redundant instances~(samples that all of its best candidates are the same) during the sampling process.} from the evaluation dataset. For each sample, the best candidate questions of $K=1$ (best predictions of PPL), $K=5$ (where the \metric\ is starting to become \textit{plateau}), and $K=50$ (best predictions of \metric) are chosen, and annotators are asked to score these three generated questions based on criteria defined in Section~\ref{sec:impl}.\footnote{Summation of scores for \textit{grammaticality}, \textit{answerability}, and \textit{relevance} is defined as the overall score.} \\
\begin{figure}
  \centering
  \includegraphics[width=\linewidth]{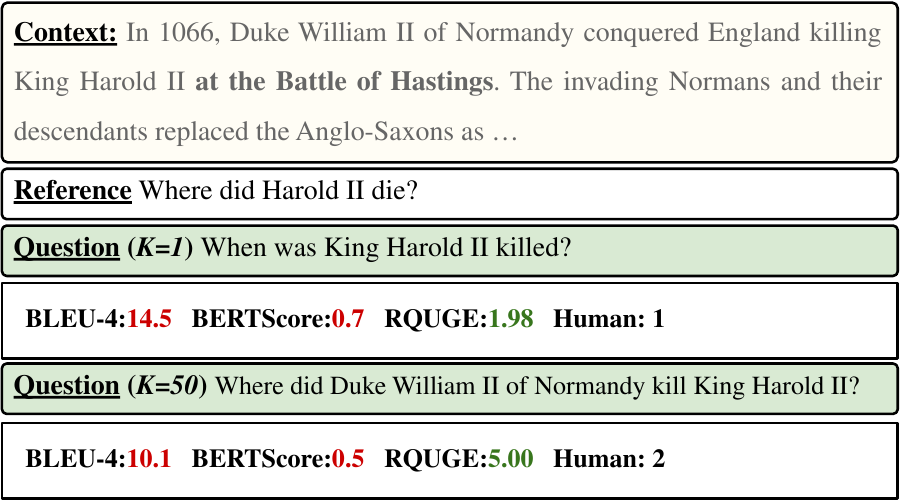}
  \caption{A sample of re-ranking experiment, that the annotator prefers the best candidate, chosen based on \metric\ ($K=50$) compared to the question selected based on the perplexity of QG model~($K=1$).}
  \label{fig:rerank_example}
\end{figure}
Figure~\ref{fig:human_rerank} depicts the relative difference of NLG metrics compared to the score of $K=1$, alongside human scores. We can see from the figure that annotators significantly prefer highest ranking questions of $K=5$ and $K=50$ chosen by \metric\ compared to the best candidate questions picked based on PPL, while the average scores of other automatic metrics drop as the number of candidate questions~($K$) increases. For $K=5$, the average human score of highest ranking questions is relatively +7.49\% better than best candidate questions, chosen by PPL of question generator~($K=1$). It confirms the effectiveness of \metric\ , when integrated into the decoding step. Additionally, there is not a significant difference~(-1.1\%) between the average human scores of candidate questions chosen based on \metric\ for $K=5$ and $K=50$. This is correlated with Figure~\ref{fig:squad_rerank}, as \metric\ also becomes \textit{plateau} from $K=5$.
Figure~\ref{fig:rerank_example} illustrates an example in which annotators prefer the second candidate question, while BLEU-4 and BERTScore compute higher scores for the first candidate question.~\footnote{For transparency, we will provide the human evaluation of the re-ranking experiment.}

\subsection{Robustness Analysis}
\label{result:robust}

To further assess the robustness of the QG metrics on adversarial corruptions of reference questions, we evaluate metrics on a subset of positive and negative samples, created from SQuAD~\cite{rajpurkar-etal-2016-squad} evaluation set, as shown in Table~\ref{tab:adv-samples}.\footnote{For a fair comparison, we omit NQ, and MS-MARCO evaluation sets as their reference questions are not always well-formed.} The remaining subset contains 2,500 samples equally selected from positive and negative questions.
\begin{table}
\centering
\begin{adjustbox}{width=\linewidth}
\begin{tabular}{lcccc}
Metric & Total & Neg. & Rev. Gen & Swap Ents \\
\toprule
\textbf{Unsupervised} \\
BLEU4 & 0.239 & 0.241 & 0.219 & 0.241 \\
ROUGE-1 & 0.148 & 0.126 & 0.209 & 0.272 \\
ROUGE-L & 0.13 & 0.11 & 0.209 & 0.272 \\
METEOR & 0.13 & 0.09 & 0.198 & 0.250 \\
QBLEU & 0.220 & 0.172 & 0.117 & 0.558 \\
MOVERScore& 0.180 & 0.169 & 0.161 & 0.236\\
BERTScore & 0.408 & 0.44 & 0.148 & 0.285 \\
\midrule 
\textbf{Regression-based} \\
BLEURT-20 & \underline{0.632} & \underline{0.69} & 0.24 & 0.489 \\
\midrule 
\textbf{Ranking-based} \\
COMET & 0.456 & 0.523 & 0.137 & 0.216 \\
\midrule 
\textbf{Generation-based} \\
BARTScore & 0.581 & 0.647 & 0.205 & 0.336 \\
\midrule 
\textbf{Ref-Free} \\
CTC & 0.539 & 0.576 & \underline{0.372} & 0.376 \\
QRelScore & 0.546 & 0.566 & \textbf{0.420} & \underline{0.535} \\
\midrule 
\textbf{\metric\ } & \textbf{0.715} & \textbf{0.759} & \underline{0.371} & \textbf{0.57} \\
\bottomrule
\end{tabular}
\end{adjustbox}
\caption{\label{tab:auc} Area under the ROC curve~(AUC) of binary classification on the adversarial subset of SQuAD dataset. First column represents the overall performance. Other columns demonstrate AUC metric for different negative sampling methods e.g. negation, reversing the gender, and swapping entities.}
\end{table}
Inspired by \citet{chen-etal-2021-factuality-checkers} and \citet{honovich-etal-2022-true}, positive questions are paraphrases of references, created by two methods, either translating to a high-resource language, then back-translating to English, or applying a T5~\cite{2020t5} model fine-tuned on Quora paraphrasing task.\footnote{\url{https://www.kaggle.com/competitions/quora-question-pairs/data}.} For negative samples, we use three strategies: negation, reversing the gender, and swapping the entities of the reference question with relevant entities in the corresponding context. Further details of the adversarial evaluation set are provided in Appendix~\ref{app:advs}. 
\begin{figure*}
    \centering
    \begin{subfigure}[b]{0.45\linewidth}        
        \centering
        \includegraphics[width=0.95\linewidth]{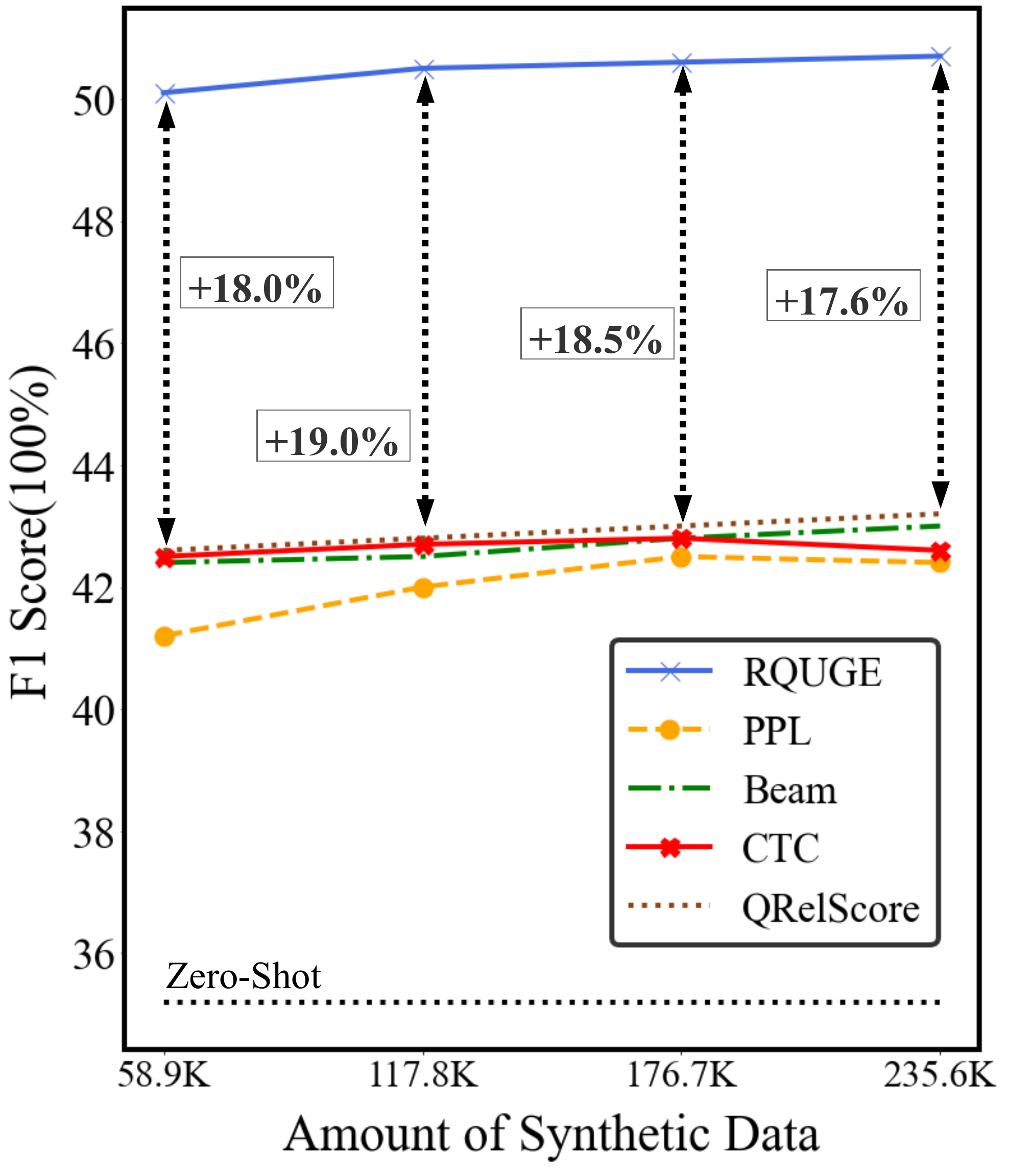}
        \caption{F1 score}
        \label{fig:A}
    \end{subfigure}
    \begin{subfigure}[b]{0.45\linewidth}        
        \centering
        \includegraphics[width=0.95\linewidth]{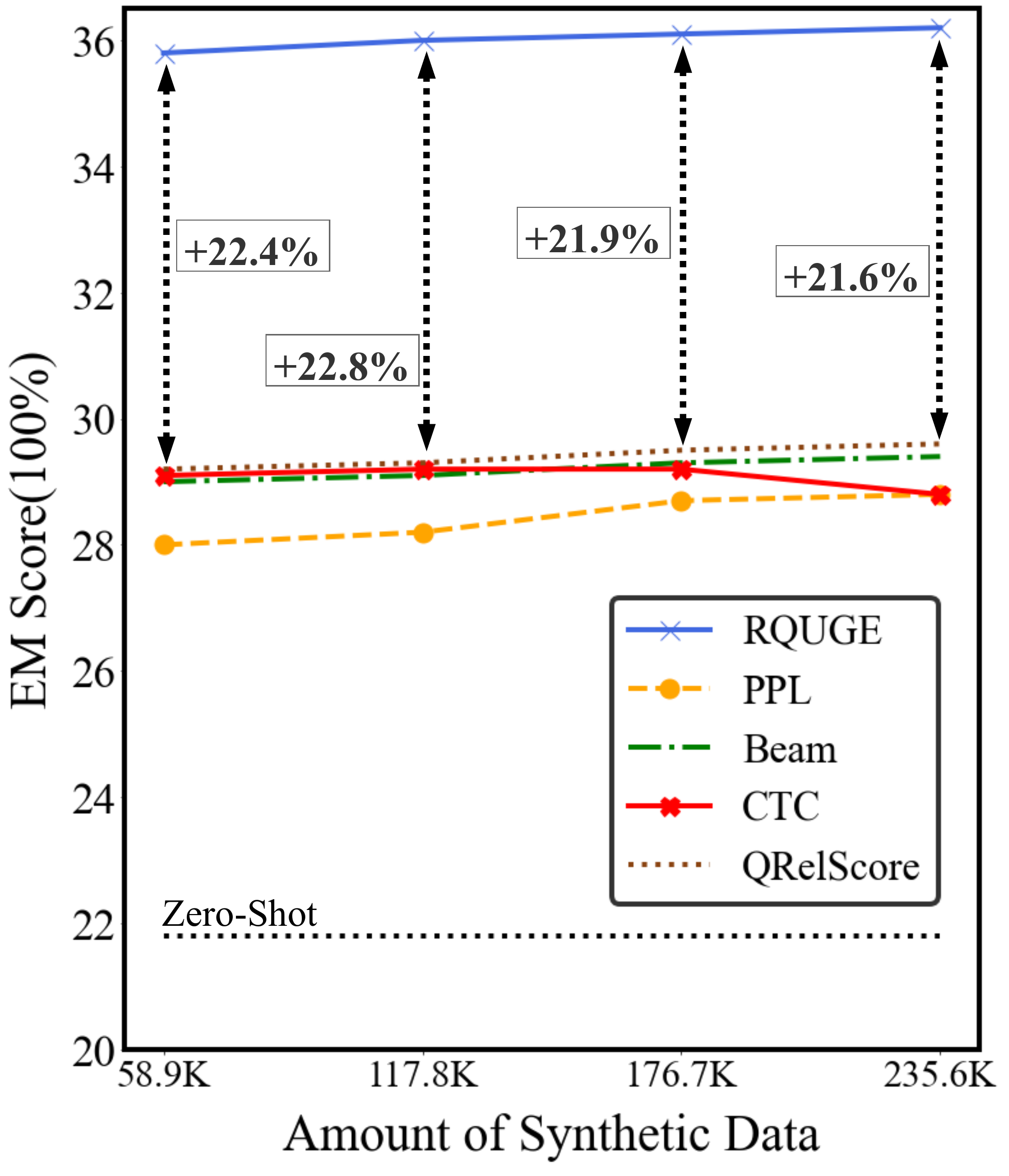}
        \caption{EM Score}
        \label{fig:B}
    \end{subfigure}
    \caption{Performance of QA model (based on F1 and Exact Match~(EM)) on out-of-domain dataset for different amount of synthetic data. Numbers in the box illustrates the relative improvement of \metric\ compared to the best baseline. 235.6K refers to the setting in which we did not see further improvement using additional synthetic data.}
    \label{fig:qa-perf}
\end{figure*}
\paragraph{Results and Discussion.} We use \metric\ and previous metrics on the adversarial subset to classify the corrupted candidate questions based on their acceptability score. Table~\ref{tab:auc} illustrates the area of the ROC curve of QG metrics on the adversarial subset.~\footnote{Number of instances for negation, gender reversing, and entity swapping are 1000, 150, and 100, respectively.} Overall, \metric\ metric significantly outperforms BLEURT-20~(the best previous work) by +13.1\% relative improvement. Previous unsupervised metrics drop significantly for all types of negative samples, while BLEURT-20, BARTScore, and reference-free metrics perform better comparatively, especially for \textit{negation}. Our \metric\ metric decreases the error relatively by +22.2\% and +7.5\% for negation, and entity swapping compared to previous work and has the second-best results on reversing the gender. This confirms the robustness of our metric for different adversarial corruption. 

\subsection{Domain Adaptation of QA Task}
\label{result:qa-task}

Generated questions using a QG model can be used to improve the performance of a question answering model on out-of-domain datasets. In this section, we show that fine-tuning on the generated synthetic data, re-ranked with \metric\ improves the performance of the question answering model. 

\paragraph{Implementation Details.} For the out-of-domain dataset, we choose MS-MARCO~\cite{msmarco2021} dataset, since the UnifiedQAv2~\cite{khashabi2022unifiedqav2} (utilised in the calculation of RQUGE) has not used it for training.~\footnote{For compatibility with the NER model, we use instances with \textit{short-form} answer span (less than 4 tokens).} Given the context, we apply Stanza~\cite{qi2020stanza} Named-Entity Recognition~(NER) model to extract candidate answer spans. A QG model is then applied to a randomly chosen candidate span, creating a bag of output predictions, using Nucleus sampling~\cite{Holtzman2020The}. Then, we apply the same re-ranking mechanism, described in Section~\ref{result-re-rank} using \metric\ , CTC, QRelScore, and the PPL of the QG. We also use a beam search of size 5~ with no re-ranking as a baseline. We use MixQG~\cite{murakhovska-etal-2022-mixqg} to generate questions. For QA, we first fine-tune T5-small~\cite{2020t5} on SQuAD~\cite{rajpurkar-etal-2016-squad} (zero-shot for our setting), then fine-tune it on the generated synthetic data. Further implementation details and hyper-parameters are provided in Appendix~\ref{app:fine-tune-qa}. 

\paragraph{Results and Discussion.} Figure~\ref{fig:qa-perf} demonstrates the performance of the QA model on the out-of-domain dataset, fine-tuned for different amounts of synthetic data. Generally, fine-tuned QA model reaches significantly better performance compared to the zero-shot setting. This is important for domains in which we do not have annotated QA data. Furthermore, fine-tuning on the re-ranked data with \metric\ consistently improves the performance of the QA model for a different amount of synthetic data, compared to other baselines. Specifically, it significantly outperforms baselines by +18.3\% F1, and +22.2\% EM, on average. It again shows the effectiveness of our \metric\ by employing it in the domain adaptation of QA models for the out-of-domain dataset.

%% file: conclusion.tex
\section{Conclusion}

We propose \metric\ , \textbf{R}eference-free \textbf{QU}estion \textbf{G}eneration \textbf{E}valuation metric to measure the quality of the generated questions, by better encoding the relevant context and answer without requiring a reference question. It consists of two modules, a question answering model, and a span scorer, which are existing pre-trained models without further fine-tuning. We compare the performance of \metric\ with existing QG metrics on SQuAD, MS-MARCO, and NQ datasets, and show that \metric\ achieves a significantly better correlation with human judgment. Additionally, we integrate \metric\ into the decoding step by using it to re-rank the candidate questions, which leads to a better correlation with human. For robustness, we evaluate QG metrics on adversarial data by corrupting the reference questions and show that \metric\ achieves significantly better performance compared to previous work. Finally, we show that fine-tuning QA models on the synthetic data, generated with a QG model and re-ranked with \metric\ , improves the performance of QA models on out-of-domain datasets.

%% file: ack.tex
\section*{Acknowledgement}

We thank Parth Pathak, Yatharf Saraf, and Omprakash Sonie for their helpful discussion and support. We are grateful to anonymous reviewers for their fruitful
comments and corrections.

%% file: limitations.tex
\section*{Limitations} The main limitation of our work is that we have applied and verified the effectiveness of our metric on the English question answering datasets. Since \metric\ depends on a strong question answering module, one has to find an alternative model to the UnifiedQA~\cite{khashabi2022unifiedqav2} we have used in calculation of \metric. Additionally, we did an error analysis on the subset that \metric\ and human evaluation have a significant difference in Appendix~\ref{app-error-analysis}, which shows that mistakes are categorised into syntactic-based and knowledge-based errors. It gives us directions for future improvement of \metric\ metric.

%% file: appendix.tex
\setcounter{section}{0}
\onecolumn
\begin{appendices}

\section{Evaluated Datsets of UnifiedQAv2 Model}
\label{app:dataset_list}

UnifiedQAv2 is evaluated on 
SQuAD~(v1)~\cite{rajpurkar-etal-2016-squad}, SQuAD~(v2)~\cite{rajpurkar-etal-2018-know}, NewsQA~\cite{trischler2016newsqa}, Quoref~\cite{Dasigi2019QuorefAR}, ROPES~\cite{Lin2019ReasoningOP},
NarrativeQA~\cite{kocisky-etal-2018-narrativeqa}, DROP~\cite{dua-etal-2019-drop}, NaturalQuestions~\cite{kwiatkowski-etal-2019-natural}, MCTest~\cite{richardson-etal-2013-mctest},
RACE~\cite{lai-etal-2017-race}, OpenBookQA~\cite{mihaylov-etal-2018-suit}, ARC~\cite{Clark2018ThinkYH}, CommonsenseQA~\cite{talmor-etal-2019-commonsenseqa},
QASC~\cite{Khot2020QASCAD}, PhysicalIQA~\cite{https://doi.org/10.48550/arxiv.1911.11641}, SocialIQA~\cite{sap-etal-2019-social}, Winogrande~\cite{10.1145/3474381},
BoolQ~\cite{clark-etal-2019-boolq}, MultiRC (yes/no)~\cite{khashabi-etal-2018-looking}, and BoolQ-NP as \textit{in-domain} datasets. Additionally, it is evaluation on AdversarialQA~\cite{advsqa2020}, ReCoRD~\cite{zhang2018record}, RACE-C~\cite{pmlr-v101-liang19a}, HeadQA~\cite{vilares-gomez-rodriguez-2019-head}, MMMLU~\cite{hendrycks2020measuring}, ReClor~\cite{Yu2020ReClor}, Quail~\cite{Rogers_Kovaleva_Downey_Rumshisky_2020}, OneStopQA~\cite{cui2021onestop}, MCScript~\cite{2018arXiv180305223O}, MCScript 2.0~\cite{ostermann-etal-2019-mcscript2}, CosmosQA~\cite{huang-etal-2019-cosmos}, DREAM~\cite{dream2019}, ProcessBank~\cite{Berant2014ModelingBP}, PROST~\cite{aroca-ouellette-etal-2021-prost}, StrategyQA~\cite{Geva2021DidAU}, PubmedQA~\cite{jin-etal-2019-pubmedqa}, QAConv~\cite{wu-etal-2022-qaconv}, and TweetQA~\cite{xiong-etal-2019-tweetqa} as \textit{out-of-domain} evaluation sets.

\section{Implementation Details}

\subsection{Details of Evaluated Datasets}
\label{app:eval_dataset}

We evaluate QG metrics on three datasets, SQuAD~(v1)~\cite{rajpurkar-etal-2016-squad}~(under CC BY-SA 4.0 license), Natural Questions~\cite{kwiatkowski-etal-2019-natural}~(under Creative Commons Share-Alike 3.0 license), and MS-MARCO~\cite{msmarco2021}~(fully open-source, no license) datasets. Table~\ref{tab:app:dataset} illustrates the number of samples in training and evaluation sets. 

\begin{table}[!ht]
\centering
\begin{adjustbox}{width=0.5\linewidth}
\begin{tabular}{lcc}
Dataset & Training Data & Evaluation Data \\
\toprule
SQuAD & 86,821 & 5,928 \\
Natural Questions & 104,071 & 12,836 \\
MS-MARCO & 502,939 & 55,578 \\
\bottomrule
\end{tabular}
\end{adjustbox}
\caption{\label{tab:app:dataset} Number of instances for the training and evaluation sets of SQuAD, \textit{short-form} of NQ, and {\tt DESCRIPTION} types of MS-MARCO datasets.}
\end{table}

\subsection{Hyper-parameters for Fine-tuning QG Models}
\label{app:hyper-finetune}
All models are trained on NVIDIA A100-SXM4-40GB GPUs. T5~\cite{2020t5} is under Apache License 2.0. GPT2~\cite{radford2019language} is under modified MIT License. We use AdamW optimiser~\cite{loshchilov2018decoupled}, used in several previous works~\cite{mohammadshahi-etal-2019-aligning,devlin-etal-2019-bert,mohammadshahi2021syntaxaware,10.1162/tacl_a_00358,mohammadshahi-henderson-2020-graph}.
\begin{table}[!ht]
\centering
\begin{minipage}[c]{0.4\textwidth}
\begin{tabular}{lcc}
Hyper-parameter & Specification \\
\toprule
Architecture & T5-base(220M) \\
No. Encoder Layers & 12 \\
No. Decoder Layers & 12 \\
No. Epochs & 15 \\
Dropout & 0.1 \\
Learning rate & 3e-5 \\
Batch size & 32 \\
No. GPUs & 8 \\
\bottomrule
\end{tabular}
\subcaption{MixQG}
\end{minipage}
\quad 
\begin{minipage}[c]{0.4\textwidth}
\begin{tabular}{lcc}
Hyper-parameter & Specification \\
\toprule
Architecture & GPT2(117M) \\
No. Encoder Layers & 12 \\
No. Epochs & 12 \\
Dropout & 0.1 \\
Learning rate & 3e-4 \\
Batch size & 32 \\
No. GPUs & 8 \\
\bottomrule
\end{tabular}
\subcaption{GPT2}
\end{minipage}
\caption{Hyper-parameters for fine-tuning QG models on evaluated datasets.}
\end{table}

\section{Instruction of Human Evaluation}
\label{app:human}

\begin{figure}[!ht]
  \centering
  \includegraphics[width=0.7\linewidth]{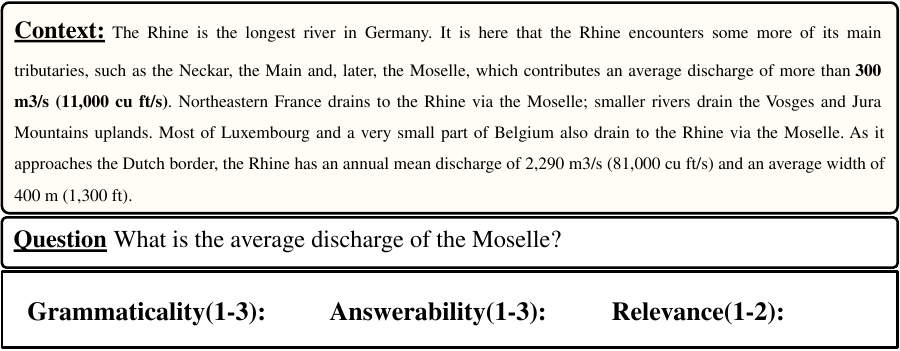}
  \caption{The input example of the human evaluation.}
  \label{app:fig:human}
\end{figure}

Annotators are asked to evaluate the quality of a question, given the context and answer span. An input example is shown in Figure~\ref{app:fig:human}. They should provide 3 scores for grammaticality, answerability, and relevance. 
For grammar, the syntactic structure of the sentence is evaluated. They should score 3 for "no grammatical errors", 2 for "not grammatically acceptable but able to get the meaning", and 1 for "unacceptable" questions. 
For answerability, the score should express the completeness of the candidate question and its consistency with the given answer. So, annotators are required to consider two criteria while scoring; the question should contain question words~(e.g. wh-words) and necessary entities, and it should not include the answer. They should score 3, if the question contains all important information, and is consistent with the answer. Score 2 is for the cases, in which some important information is missing in the question or it contains the answer. They should score 1 if all important information is missing in the question and the question is not consistent with the answer.\\
For relevance, annotators should score the relatedness of the question to the answer, given the context. They should score 2 if the question is answerable by the context and related to the given answer. They should score 1, if the question is out-of-context, or not related to the given answer. \\
We sample 600 examples (200 for each dataset) from the evaluation sets of SQuAD~(v1), NQ, and MS-MARCO. Samples are shuffled and anonymized. All annotators are fluent English speakers.

\section{Correlation with Human Evaluation}
\label{app:result_human}
~~
\begin{table}[!htb]
\centering
  \begin{adjustbox}{width=0.5\linewidth}
  \begin{tabular}{lccccccccc}
    \toprule
    \multirow{2}{*}{Metric} & 
      \multicolumn{3}{c}{Answerability} &&
      \multicolumn{3}{c}{Relevance} \\
      \cline{2-4} \cline{6-8} 
     & $r$ & $\rho$ & $\tau$ && $r$ & $\rho$ & $\tau$ \\
    \midrule
    \textbf{Unsupervised} &    &  &  &\vline&   &  & & \\
    BLEU-4 & 0.256 & 0.291 & 0.224  &\vline& 0.213 & 0.207 & 0.168\\
    ROUGE-1 & 0.317 & 0.292 & 0.230  &\vline& 0.303 & 0.281 & 0.231 \\
    ROUGE-L  & 0.345 & 0.332 & 0.263 &\vline&  0.312 & 0.286 & 0.235 \\
    METEOR  & 0.337 & 0.316 & 0.249  &\vline& 0.308 & 0.291 & 0.237 \\
    QBLEU  & 0.296 & 0.300 & 0.238 &\vline& 0.272 & 0.274 & 0.223 \\
    MOVERScore  & 0.296 & 0.317 & 0.248 &\vline&  0.258 & 0.270 & 0.218 \\
    BERTScore  & 0.344 & 0.343 & 0.266  &\vline& 0.306 & 0.288 & 0.233 \\
    \midrule
    \textbf{Regression-based}  &    &  &  &\vline&   &  & & \\
    BLEURT-20  & 0.340 & 0.311 & 0.242 &\vline&  0.326 & 0.306 & 0.247 \\
    \midrule
    \textbf{Ranking-based}  &    &  &  &\vline&   &  & & \\
    COMET  & 0.355 & 0.359 & 0.279  &\vline& 0.271 & 0.276 & 0.222 \\
    \midrule
    \textbf{Generation-based}  &    &  &  &\vline&   &  & & \\
    BARTScore  & \underline{0.391} & \underline{0.383} & \underline{0.300} &\vline&  \underline{0.378} & \underline{0.334} & \underline{0.270} \\
    \midrule 
    \textbf{Ref-Free}  &    &  &  &\vline&   &  & & \\
    CTC  &  0.236 & 0.130 & 0.099  &\vline&  0.271 & 0.189 & 0.152 \\
    QRelScore  &  0.332 & 0.276 & 0.212  &\vline&  0.262 & 0.213 & 0.175 \\
    \midrule 
    \textbf{\metric }  & \textbf{0.688} & \textbf{0.388} & \textbf{0.303} &\vline& \textbf{0.588} & \textbf{0.404} & \textbf{0.327} \\
    \bottomrule
  \end{tabular}
  \end{adjustbox}
  \caption{Correlation of human judgment and evaluation metrics based on Pearson $r$, Spearman $\rho$, and Kendall $\tau$ correlation coefficients on SQuAD~(v1)~\cite{rajpurkar-etal-2016-squad} dataset.}
      \vspace{-1ex}
\end{table}
~~
\begin{table}[!htb]
\centering
  \begin{adjustbox}{width=0.5\linewidth}
  \begin{tabular}{lccccccccc}
    \toprule
    \multirow{2}{*}{Metric} & 
      \multicolumn{3}{c}{Answerability} &&
      \multicolumn{3}{c}{Relevance} \\
      \cline{2-4} \cline{6-8} 
     & $r$ & $\rho$ & $\tau$ && $r$ & $\rho$ & $\tau$ \\

    \midrule
    \textbf{Unsupervised} &    &  &  &\vline&   &  & & \\
    BLEU-4 & 0.405 & 0.494 & 0.398  &\vline& 0.393 & 0.467 & 0.380 \\
    ROUGE-1 & 0.533 & 0.530 & 0.430  &\vline& \underline{0.517} & \underline{0.510} & \underline{0.418} \\
    ROUGE-L  & 0.514 & 0.523 & 0.425 &\vline& 0.491 & 0.491 & 0.403 \\
    METEOR  & \underline{0.533} & \underline{0.535} & \underline{0.430}  &\vline& 0.513 & 0.505 & 0.411 \\
    QBLEU  & 0.502 & 0.524 & 0.417 &\vline& 0.500 & 0.497 & 0.405 \\
    MOVERScore  & 0.434 & 0.482 & 0.385 &\vline& 0.419 & 0.480 & 0.391 \\
    BERTScore  & 0.480 & 0.490 & 0.398 &\vline& 0.488 & 0.495 & 0.406 \\
    \midrule
    \textbf{Regression-based}  &    &  &  &\vline&   &  & & \\
    BLEURT-20  & 0.501 & 0.506 & 0.408 &\vline&  0.488 & 0.509 & 0.413 \\
    \midrule
    \textbf{Ranking-based}  &    &  &  &\vline&   &  & & \\
    COMET  & 0.405 & 0.393 & 0.310  &\vline&  0.397 & 0.403 & 0.325 \\
    \midrule
    \textbf{Generation-based}  &    &  &  &\vline&   &  & & \\
    BARTScore  & 0.421 & 0.439 & 0.351  &\vline& 0.419 & 0.437 & 0.358 \\
    \midrule 
    \textbf{Ref-Free}  &    &  &  &\vline&   &  & & \\
    CTC  &  0.270 & 0.266 & 0.208 &\vline& 0.270 & 0.254 & 0.207  \\
    QRelScore  &  0.415 & 0.309 & 0.276 &\vline& 0.394 & 0.292 & 0.264  \\
    \midrule 
    \textbf{\metric } &  \textbf{0.781} & \textbf{0.564} & \textbf{0.446} &\vline&  \textbf{0.783} & \textbf{0.592} & \textbf{0.476} \\
    \bottomrule
  \end{tabular}
  \end{adjustbox}
  \caption{Correlation of human judgment and evaluation metrics based on Pearson $r$, Spearman $\rho$, and Kendall $\tau$ correlation coefficients on Natural Questions~\cite{kwiatkowski-etal-2019-natural} dataset.}
      \vspace{-1ex}
\end{table}
~~
\begin{table}[!htb]
\centering
  \begin{adjustbox}{width=0.5\linewidth}
  \begin{tabular}{lccccccccc}
    \toprule
    \multirow{2}{*}{Metric} & 
      \multicolumn{3}{c}{Answerability} &&
      \multicolumn{3}{c}{Relevance} \\
      \cline{2-4} \cline{6-8} 
     & $r$ & $\rho$ & $\tau$ && $r$ & $\rho$ & $\tau$ \\
    \midrule
    \textbf{Unsupervised} &    &  &  &\vline&   &  & \\
    BLEU-4 & 0.222 & 0.272 & 0.211  &\vline& 0.096 & 0.109 & 0.089 \\
    ROUGE-1 & 0.107 & 0.086 & 0.070  &\vline& 0.174 & 0.199 & 0.165\\
    ROUGE-L & 0.146 & 0.131 & 0.106 &\vline& 0.180 & 0.200 & 0.165 \\
    METEOR  & 0.168 & 0.167 & 0.131  &\vline& 0.167 & 0.181 & 0.145 \\
    QBLEU  &  0.138 & 0.128 & 0.101  &\vline&  0.134 & 0.134 & 0.108 \\
    MOVERScore  & 0.200 & 0.217 & 0.168 &\vline& 0.197 & 0.206 & 0.167 \\
    BERTScore  &  0.201 & 0.205 & 0.159  &\vline&  0.153 & 0.165 & 0.134 \\
    \midrule
    \textbf{Regression-based}  &    &  &  &\vline&   &  & \\
    BLEURT-20  & 0.246 & 0.255 & 0.202  &\vline&  \underline{0.275} & \underline{0.280} & \underline{0.229} \\
    \midrule
    \textbf{Ranking-based}  &    &  &  &\vline&   &  & \\
    COMET  & 0.209 & 0.229 & 0.181 &\vline&  0.244 & 0.261 & 0.213 \\
    \midrule 
    \textbf{Generation-based}  &    &  &  &\vline&   &  & & \\
    BARTScore  & 0.221 & 0.236 & 0.184 &\vline& 0.181 & 0.207 & 0.168 \\
    \midrule
    \textbf{Ref-Free}  &    &  &  &\vline&   &  & & \\
    CTC  & \textbf{0.401} & \textbf{0.376} & \underline{0.290}  &\vline& 0.154 & 0.183 & 0.150  \\
    QRelScore  & 0.351 & 0.272 & 0.172  &\vline& 0.226 & 0.133 & 0.126  \\
    \midrule 
    \textbf{\metric\ } & \underline{0.400} & \underline{0.366} & \textbf{0.293}  &\vline& \textbf{0.397} & \textbf{0.356} & \textbf{0.288} \\
    \bottomrule
  \end{tabular}
  \end{adjustbox}
  \caption{Correlation of human judgment and evaluation metrics based on Pearson $r$, Spearman $\rho$, and Kendall $\tau$ correlation coefficients on MS-MARCO~\cite{msmarco2021} dataset.}
      \vspace{-1ex}
\end{table}
~~

~
\newpage
~

\section{Re-Ranking with RQUGE}
\label{app:reranking}

\begin{figure}[!ht]
  \centering
  \includegraphics[width=0.6\linewidth]{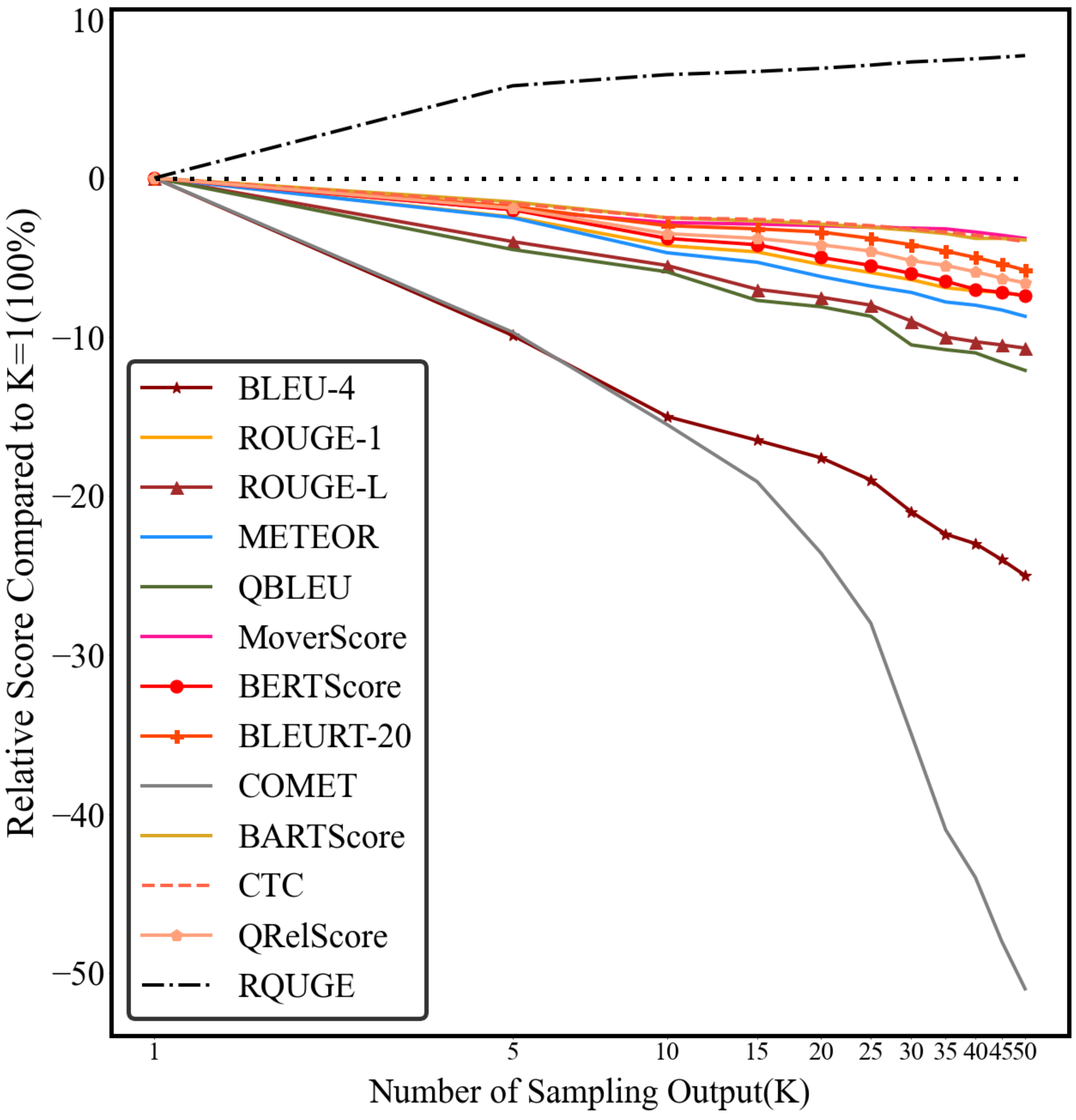}
  \caption{The relative score of automatic metrics compared to $K=1$ for different values of $K$, after re-ranking the output of question generation model with \metric\ metric. \metric\ increases as it is the objective of the re-ranking mechanism.}
  \label{app:fig:squad_rerank}
\end{figure}

\section{Adversarial Evaluation Set}
\label{app:advs}

As discussed in Section~\ref{result:robust}, we create positive samples by two mechanisms:
\begin{itemize}
\item \textbf{Back-Translation.} Translating the reference question to an intermediate language, then translating it back to English. We apply Marian model~\cite{junczys-dowmunt-etal-2018-marian}, and use Chinese and French as intermediate languages, as Marian model has reasonable performance for these language directions. 
\item \textbf{Quora Paraphrasing.} We first train a T5-small~\cite{2020t5} model on Quora paraphrasing dataset,~\footnote{\url{https://www.kaggle.com/competitions/quora-question-pairs/data}} and use it for paraphrasing the reference question. 
\end{itemize}
Outputs of both methods are questions that are semantically similar to the reference questions with a few lexical differences. \\
For the negative samples, as shown in Table~\ref{tab:adv-samples}, we apply the following methods:
\begin{itemize}
    \item \textbf{Negation.} We first scan the reference question to find auxiliary and modal verbs. Then, we randomly either add \textit{not} to the sentence or replace the verb with its antonyms by using WordNet~\cite{10.1145/219717.219748} inside the NLTK package~\cite{bird2009natural}.
    \item \textbf{Reverse Gender.} The reference question is first scanned to find pronouns, and then pronouns are replaced with pronouns with the opposite gender. 
    \item \textbf{Swap Entity.} Stanza~\cite{qi-etal-2020-stanza} named-entity recognition model is applied to the reference question and the context. Then, we randomly select one extracted entity of the reference question and replace it with a random entity of the context with the same entity type. 
\end{itemize}

\section{Implementation Details of Fine-tuning QA models}
\label{app:fine-tune-qa}
All models are trained on NVIDIA A100-SXM4-40GB GPUs.

\begin{table}[!htb]
\centering
\begin{adjustbox}{width=0.3\linewidth}
\begin{tabular}{lcc}
Hyper-parameter & Specification \\
\toprule
Architecture & T5-small \\
No. Encoder Layers & 6 \\
No. Decoder Layers & 6 \\
No. Training Steps & 2K \\
Dropout & 0.1 \\
Learning rate & 3e-5 \\
Batch size & 32 \\
No. GPUs & 8 \\
\bottomrule
\end{tabular}
\end{adjustbox}
\caption{Hyper-parameters for fine-tuning QA models on the synthetic data of MS-MARCO.}
\end{table}

\section{Error Analysis}
\label{app-error-analysis}

\newcolumntype{L}{>{\raggedright\arraybackslash}X}

We investigate on cases, in which there is a substantial difference between the human evaluation and \metric\ score. The errors are categorised into syntactic and knowledge-based types, as shown in Table~\ref{app:error-tab1}. For the syntactic error, \metric\ sometimes computes unacceptable scores for sentences that either miss the question word~(e.g. wh-words) or have wrong word order, as QA module of \metric\ focuses more on the semantic aspect of the candidate question to predict the answer span. For the knowledge-based mistakes, \metric\ requires further domain specific and commonsense knowledge to compute the correct score e.g. full moon is instant, not period of time as illustrated in the sample of Table~\ref{app:error-tab1}. As shown in Table~\ref{app:error_tab2}, \metric\ computes wrong values for some samples in "reversing gender" and "swapping entities" categories of evaluation set in Section~\ref{result:robust}. \\
These errors shows the limitations of \metric\ metric, and lead the future work to apply larger and better QA and span scorer modules. 

\begin{table}[!htb]
\centering
\begin{adjustbox}{width=0.9\linewidth}
\begin{tabularx}{\linewidth}{LLLcc}
\small 
Error Type & \small Question & \small Answer \& Context & \small \metric\ & \small Avg Human \\
\toprule
\small{\textbf{Syntactic}} \\
\small{(1)} & \small{cost of wooden shutters} &  \small{Exterior window shutters cover ... Typical costs: Wooden or vinyl exterior window shutters in stock sizes cost \textbf{\$20-\$200 per pair of panels}.} & \small{4.81/5} & \small{grammaticality:1.66/3} \\
 \small{(2)} & \small{SAT solvers routinely handle large instances of what?} & \small{... Similarly, algorithms can solve the NP-complete knapsack problem over a wide range of sizes in less than quadratic time and SAT solvers routinely handle large instances of the \textbf{NP-complete Boolean satisfiability problem}.} & \small{4.76/5} & \small{grammaticality:2/3} \\ \hline 
 \small{\textbf{Knowledge-based}} & \small{how long is a full moon} & \small{\textbf{A full lunar cycle lasts almost a month (about 29.5 days)}, and ... However, a full moon, a new moon, and a half moon (first and third quarter) are instants, not periods of time.} & \small{4.95/5} & \small{answerability:1/3, relevance:1/3} \\
\bottomrule
\end{tabularx}
\end{adjustbox}
\caption{Different categories of errors that \metric\ metric computes wrong scores. \label{app:error-tab1}}
\end{table}

\begin{table}[!htb]
\centering
\begin{adjustbox}{width=0.9\linewidth}
\begin{tabularx}{\linewidth}{LLLLc}
\small Corruption Type & \small{Ref Question} & \small Corrupted Question & \small Answer \& Context & \small \metric\ \\
\toprule
\small{Reversing gender} & \small{In what year was the university's 5th president granted his position?} & \small{In what year was the university's 5th president granted hers position?} & \small{In \textbf{1929}, the university's fifth president, Robert Maynard Hutchins, took office; the university underwent many changes during his 24-year tenure...} & \small{2.25/5} \\ \hline 
\small{Swapping entities} & \small{The Kuznets curve says with economic development, inequality will decrease after what?} & \small{The Piketty curve says with economic development, inequality will decrease after what?} & \small{Studies on income inequality and growth have sometimes found evidence confirming the Kuznets curve hypothesis, which states that with economic development, \textbf{inequality first increases}, then decreases. Economist Thomas Piketty challenges this notion...} & \small{4.65/5} \\
\bottomrule
\end{tabularx}
\end{adjustbox}
\caption{Some samples from the adversarial subset of Section~\ref{result:robust}, that \metric\ metric is not sensitive to the corruption. \label{app:error_tab2}}
\end{table}

\end{appendices}

%% file: main.bbl
\begin{thebibliography}{99}
\expandafter\ifx\csname natexlab\endcsname\relax\def\natexlab#1{#1}\fi

\bibitem[{Aroca-Ouellette et~al.(2021)Aroca-Ouellette, Paik, Roncone, and
  Kann}]{aroca-ouellette-etal-2021-prost}
St{\'e}phane Aroca-Ouellette, Cory Paik, Alessandro Roncone, and Katharina
  Kann. 2021.
\newblock \href {https://doi.org/10.18653/v1/2021.findings-acl.404} {{PROST}:
  {P}hysical reasoning about objects through space and time}.
\newblock In \emph{Findings of the Association for Computational Linguistics:
  ACL-IJCNLP 2021}, pages 4597--4608, Online. Association for Computational
  Linguistics.

\bibitem[{Bartolo et~al.(2020)Bartolo, Roberts, Welbl, Riedel, and
  Stenetorp}]{advsqa2020}
Max Bartolo, Alastair Roberts, Johannes Welbl, Sebastian Riedel, and Pontus
  Stenetorp. 2020.
\newblock \href {https://doi.org/10.1162/tacl_a_00338} {Beat the ai:
  Investigating adversarial human annotation for reading comprehension}.
\newblock \emph{Transactions of the Association for Computational Linguistics},
  8:662–678.

\bibitem[{Berant et~al.(2014)Berant, Srikumar, Chen, Linden, Harding, Huang,
  Clark, and Manning}]{Berant2014ModelingBP}
Jonathan Berant, Vivek Srikumar, Pei-Chun Chen, Abby~Vander Linden, Brittany
  Harding, Brad Huang, Peter Clark, and Christopher~D. Manning. 2014.
\newblock Modeling biological processes for reading comprehension.
\newblock In \emph{EMNLP}.

\bibitem[{Bird et~al.(2009)Bird, Klein, and Loper}]{bird2009natural}
Steven Bird, Ewan Klein, and Edward Loper. 2009.
\newblock \emph{Natural language processing with Python: analyzing text with
  the natural language toolkit}.
\newblock " O'Reilly Media, Inc.".

\bibitem[{Bisk et~al.(2019)Bisk, Zellers, Bras, Gao, and
  Choi}]{https://doi.org/10.48550/arxiv.1911.11641}
Yonatan Bisk, Rowan Zellers, Ronan~Le Bras, Jianfeng Gao, and Yejin Choi. 2019.
\newblock \href {https://doi.org/10.48550/ARXIV.1911.11641} {Piqa: Reasoning
  about physical commonsense in natural language}.

\bibitem[{Bonifacio et~al.(2021)Bonifacio, Jeronymo, Abonizio, Campiotti,
  Fadaee, Lotufo, and Nogueira}]{msmarco2021}
Luiz Bonifacio, Vitor Jeronymo, Hugo~Queiroz Abonizio, Israel Campiotti,
  Marzieh Fadaee, Roberto Lotufo, and Rodrigo Nogueira. 2021.
\newblock \href {https://doi.org/10.48550/ARXIV.2108.13897} {mmarco: A
  multilingual version of the ms marco passage ranking dataset}.

\bibitem[{Chen et~al.(2020)Chen, Stanovsky, Singh, and
  Gardner}]{chen-etal-2020-mocha}
Anthony Chen, Gabriel Stanovsky, Sameer Singh, and Matt Gardner. 2020.
\newblock \href {https://doi.org/10.18653/v1/2020.emnlp-main.528} {{MOCHA}: A
  dataset for training and evaluating generative reading comprehension
  metrics}.
\newblock In \emph{Proceedings of the 2020 Conference on Empirical Methods in
  Natural Language Processing (EMNLP)}, pages 6521--6532, Online. Association
  for Computational Linguistics.

\bibitem[{Chen et~al.(2018)Chen, Yang, Hauff, and
  Houben}]{Chen_Yang_Hauff_Houben_2018}
Guanliang Chen, Jie Yang, Claudia Hauff, and Geert-Jan Houben. 2018.
\newblock \href {https://ojs.aaai.org/index.php/ICWSM/article/view/14987}
  {Learningq: A large-scale dataset for educational question generation}.
\newblock \emph{Proceedings of the International AAAI Conference on Web and
  Social Media}, 12(1).

\bibitem[{Chen et~al.(2021)Chen, Liu, and
  Qiu}]{chen-etal-2021-factuality-checkers}
Yiran Chen, Pengfei Liu, and Xipeng Qiu. 2021.
\newblock \href {https://doi.org/10.18653/v1/2021.findings-emnlp.179} {Are
  factuality checkers reliable? adversarial meta-evaluation of factuality in
  summarization}.
\newblock In \emph{Findings of the Association for Computational Linguistics:
  EMNLP 2021}, pages 2082--2095, Punta Cana, Dominican Republic. Association
  for Computational Linguistics.

\bibitem[{Cheng et~al.(2021)Cheng, Li, Liu, Zhao, Li, Lin, and
  Zheng}]{cheng-etal-2021-guiding}
Yi~Cheng, Siyao Li, Bang Liu, Ruihui Zhao, Sujian Li, Chenghua Lin, and Yefeng
  Zheng. 2021.
\newblock \href {https://doi.org/10.18653/v1/2021.acl-long.465} {Guiding the
  growth: Difficulty-controllable question generation through step-by-step
  rewriting}.
\newblock In \emph{Proceedings of the 59th Annual Meeting of the Association
  for Computational Linguistics and the 11th International Joint Conference on
  Natural Language Processing (Volume 1: Long Papers)}, pages 5968--5978,
  Online. Association for Computational Linguistics.

\bibitem[{Clark et~al.(2019)Clark, Lee, Chang, Kwiatkowski, Collins, and
  Toutanova}]{clark-etal-2019-boolq}
Christopher Clark, Kenton Lee, Ming-Wei Chang, Tom Kwiatkowski, Michael
  Collins, and Kristina Toutanova. 2019.
\newblock \href {https://doi.org/10.18653/v1/N19-1300} {{B}ool{Q}: Exploring
  the surprising difficulty of natural yes/no questions}.
\newblock In \emph{Proceedings of the 2019 Conference of the North {A}merican
  Chapter of the Association for Computational Linguistics: Human Language
  Technologies, Volume 1 (Long and Short Papers)}, pages 2924--2936,
  Minneapolis, Minnesota. Association for Computational Linguistics.

\bibitem[{Clark et~al.(2018)Clark, Cowhey, Etzioni, Khot, Sabharwal, Schoenick,
  and Tafjord}]{Clark2018ThinkYH}
Peter Clark, Isaac Cowhey, Oren Etzioni, Tushar Khot, Ashish Sabharwal, Carissa
  Schoenick, and Oyvind Tafjord. 2018.
\newblock Think you have solved question answering? try arc, the ai2 reasoning
  challenge.
\newblock \emph{ArXiv}, abs/1803.05457.

\bibitem[{Cui et~al.(2021)Cui, Bao, Zu, Guo, Zhao, Zhang, and
  Chen}]{cui2021onestop}
Shaobo Cui, Xintong Bao, Xinxing Zu, Yangyang Guo, Zhongzhou Zhao, Ji~Zhang,
  and Haiqing Chen. 2021.
\newblock \href {http://arxiv.org/abs/2102.12128} {Onestop qamaker: Extract
  question-answer pairs from text in a one-stop approach}.

\bibitem[{Dasigi et~al.(2019)Dasigi, Liu, Marasovi{\'c}, Smith, and
  Gardner}]{Dasigi2019QuorefAR}
Pradeep Dasigi, Nelson~F. Liu, Ana Marasovi{\'c}, Noah~A. Smith, and Matt
  Gardner. 2019.
\newblock Quoref: A reading comprehension dataset with questions requiring
  coreferential reasoning.
\newblock In \emph{EMNLP}.

\bibitem[{Deng et~al.(2021)Deng, Tan, Liu, Xing, and
  Hu}]{deng-etal-2021-compression}
Mingkai Deng, Bowen Tan, Zhengzhong Liu, Eric Xing, and Zhiting Hu. 2021.
\newblock \href {https://doi.org/10.18653/v1/2021.emnlp-main.599} {Compression,
  transduction, and creation: A unified framework for evaluating natural
  language generation}.
\newblock In \emph{Proceedings of the 2021 Conference on Empirical Methods in
  Natural Language Processing}, pages 7580--7605, Online and Punta Cana,
  Dominican Republic. Association for Computational Linguistics.

\bibitem[{Denkowski and Lavie(2010)}]{denkowski-lavie-2010-extending}
Michael Denkowski and Alon Lavie. 2010.
\newblock \href {https://aclanthology.org/N10-1031} {Extending the {METEOR}
  machine translation evaluation metric to the phrase level}.
\newblock In \emph{Human Language Technologies: The 2010 Annual Conference of
  the North {A}merican Chapter of the Association for Computational
  Linguistics}, pages 250--253, Los Angeles, California. Association for
  Computational Linguistics.

\bibitem[{Devlin et~al.(2019)Devlin, Chang, Lee, and
  Toutanova}]{devlin-etal-2019-bert}
Jacob Devlin, Ming-Wei Chang, Kenton Lee, and Kristina Toutanova. 2019.
\newblock \href {https://doi.org/10.18653/v1/N19-1423} {{BERT}: Pre-training of
  deep bidirectional transformers for language understanding}.
\newblock In \emph{Proceedings of the 2019 Conference of the North {A}merican
  Chapter of the Association for Computational Linguistics: Human Language
  Technologies, Volume 1 (Long and Short Papers)}, pages 4171--4186,
  Minneapolis, Minnesota. Association for Computational Linguistics.

\bibitem[{Du and Cardie(2018)}]{du-cardie-2018-harvesting}
Xinya Du and Claire Cardie. 2018.
\newblock \href {https://doi.org/10.18653/v1/P18-1177} {Harvesting
  paragraph-level question-answer pairs from {W}ikipedia}.
\newblock In \emph{Proceedings of the 56th Annual Meeting of the Association
  for Computational Linguistics (Volume 1: Long Papers)}, pages 1907--1917,
  Melbourne, Australia. Association for Computational Linguistics.

\bibitem[{Du et~al.(2017)Du, Shao, and Cardie}]{du-etal-2017-learning}
Xinya Du, Junru Shao, and Claire Cardie. 2017.
\newblock \href {https://doi.org/10.18653/v1/P17-1123} {Learning to ask: Neural
  question generation for reading comprehension}.
\newblock In \emph{Proceedings of the 55th Annual Meeting of the Association
  for Computational Linguistics (Volume 1: Long Papers)}, pages 1342--1352,
  Vancouver, Canada. Association for Computational Linguistics.

\bibitem[{Dua et~al.(2019)Dua, Wang, Dasigi, Stanovsky, Singh, and
  Gardner}]{dua-etal-2019-drop}
Dheeru Dua, Yizhong Wang, Pradeep Dasigi, Gabriel Stanovsky, Sameer Singh, and
  Matt Gardner. 2019.
\newblock \href {https://doi.org/10.18653/v1/N19-1246} {{DROP}: A reading
  comprehension benchmark requiring discrete reasoning over paragraphs}.
\newblock In \emph{Proceedings of the 2019 Conference of the North {A}merican
  Chapter of the Association for Computational Linguistics: Human Language
  Technologies, Volume 1 (Long and Short Papers)}, pages 2368--2378,
  Minneapolis, Minnesota. Association for Computational Linguistics.

\bibitem[{Duan et~al.(2017)Duan, Tang, Chen, and
  Zhou}]{duan-etal-2017-question}
Nan Duan, Duyu Tang, Peng Chen, and Ming Zhou. 2017.
\newblock \href {https://doi.org/10.18653/v1/D17-1090} {Question generation for
  question answering}.
\newblock In \emph{Proceedings of the 2017 Conference on Empirical Methods in
  Natural Language Processing}, pages 866--874, Copenhagen, Denmark.
  Association for Computational Linguistics.

\bibitem[{Fabbri et~al.(2022)Fabbri, Wu, Liu, and
  Xiong}]{fabbri-etal-2022-qafacteval}
Alexander Fabbri, Chien-Sheng Wu, Wenhao Liu, and Caiming Xiong. 2022.
\newblock \href {https://doi.org/10.18653/v1/2022.naacl-main.187}
  {{QAF}act{E}val: Improved {QA}-based factual consistency evaluation for
  summarization}.
\newblock In \emph{Proceedings of the 2022 Conference of the North American
  Chapter of the Association for Computational Linguistics: Human Language
  Technologies}, pages 2587--2601, Seattle, United States. Association for
  Computational Linguistics.

\bibitem[{Geva et~al.(2021)Geva, Khashabi, Segal, Khot, Roth, and
  Berant}]{Geva2021DidAU}
Mor Geva, Daniel Khashabi, Elad Segal, Tushar Khot, Dan Roth, and Jonathan
  Berant. 2021.
\newblock Did aristotle use a laptop? a question answering benchmark with
  implicit reasoning strategies.
\newblock \emph{Transactions of the Association for Computational Linguistics},
  9:346--361.

\bibitem[{He et~al.(2021)He, Liu, Gao, and Chen}]{he2021deberta}
Pengcheng He, Xiaodong Liu, Jianfeng Gao, and Wei Chen. 2021.
\newblock \href
  {https://www.microsoft.com/en-us/research/publication/deberta-decoding-enhanced-bert-with-disentangled-attention-2/}
  {Deberta: Decoding-enhanced bert with disentangled attention}.
\newblock In \emph{2021 International Conference on Learning Representations}.
\newblock Under review.

\bibitem[{Hendrycks et~al.(2020)Hendrycks, Burns, Basart, Zou, Mazeika, Song,
  and Steinhardt}]{hendrycks2020measuring}
Dan Hendrycks, Collin Burns, Steven Basart, Andy Zou, Mantas Mazeika, Dawn
  Song, and Jacob Steinhardt. 2020.
\newblock \href {http://arxiv.org/abs/2009.03300} {Measuring massive multitask
  language understanding}.

\bibitem[{Hirao et~al.(2007)Hirao, Okumura, Yasuda, and
  Isozaki}]{HIRAO20071521}
Tsutomu Hirao, Manabu Okumura, Norihito Yasuda, and Hideki Isozaki. 2007.
\newblock \href {https://doi.org/https://doi.org/10.1016/j.ipm.2007.01.012}
  {Supervised automatic evaluation for summarization with voted regression
  model}.
\newblock \emph{Information Processing \& Management}, 43(6):1521--1535.
\newblock Text Summarization.

\bibitem[{Holtzman et~al.(2020)Holtzman, Buys, Du, Forbes, and
  Choi}]{Holtzman2020The}
Ari Holtzman, Jan Buys, Li~Du, Maxwell Forbes, and Yejin Choi. 2020.
\newblock \href {https://openreview.net/forum?id=rygGQyrFvH} {The curious case
  of neural text degeneration}.
\newblock In \emph{International Conference on Learning Representations}.

\bibitem[{Honovich et~al.(2022)Honovich, Aharoni, Herzig, Taitelbaum,
  Kukliansy, Cohen, Scialom, Szpektor, Hassidim, and
  Matias}]{honovich-etal-2022-true}
Or~Honovich, Roee Aharoni, Jonathan Herzig, Hagai Taitelbaum, Doron Kukliansy,
  Vered Cohen, Thomas Scialom, Idan Szpektor, Avinatan Hassidim, and Yossi
  Matias. 2022.
\newblock \href {https://doi.org/10.18653/v1/2022.dialdoc-1.19} {{TRUE}:
  Re-evaluating factual consistency evaluation}.
\newblock In \emph{Proceedings of the Second DialDoc Workshop on
  Document-grounded Dialogue and Conversational Question Answering}, pages
  161--175, Dublin, Ireland. Association for Computational Linguistics.

\bibitem[{Hosking and Riedel(2019)}]{hosking-riedel-2019-evaluating}
Tom Hosking and Sebastian Riedel. 2019.
\newblock \href {https://doi.org/10.18653/v1/N19-1237} {Evaluating rewards for
  question generation models}.
\newblock In \emph{Proceedings of the 2019 Conference of the North {A}merican
  Chapter of the Association for Computational Linguistics: Human Language
  Technologies, Volume 1 (Long and Short Papers)}, pages 2278--2283,
  Minneapolis, Minnesota. Association for Computational Linguistics.

\bibitem[{Hu et~al.(2019)Hu, Singh, Holzenberger, Post, and
  Van~Durme}]{hu-etal-2019-large}
J.~Edward Hu, Abhinav Singh, Nils Holzenberger, Matt Post, and Benjamin
  Van~Durme. 2019.
\newblock \href {https://doi.org/10.18653/v1/K19-1005} {Large-scale, diverse,
  paraphrastic bitexts via sampling and clustering}.
\newblock In \emph{Proceedings of the 23rd Conference on Computational Natural
  Language Learning (CoNLL)}, pages 44--54, Hong Kong, China. Association for
  Computational Linguistics.

\bibitem[{Huang et~al.(2019)Huang, Le~Bras, Bhagavatula, and
  Choi}]{huang-etal-2019-cosmos}
Lifu Huang, Ronan Le~Bras, Chandra Bhagavatula, and Yejin Choi. 2019.
\newblock \href {https://doi.org/10.18653/v1/D19-1243} {Cosmos {QA}: Machine
  reading comprehension with contextual commonsense reasoning}.
\newblock In \emph{Proceedings of the 2019 Conference on Empirical Methods in
  Natural Language Processing and the 9th International Joint Conference on
  Natural Language Processing (EMNLP-IJCNLP)}, pages 2391--2401, Hong Kong,
  China. Association for Computational Linguistics.

\bibitem[{Jin et~al.(2019)Jin, Dhingra, Liu, Cohen, and
  Lu}]{jin-etal-2019-pubmedqa}
Qiao Jin, Bhuwan Dhingra, Zhengping Liu, William Cohen, and Xinghua Lu. 2019.
\newblock \href {https://doi.org/10.18653/v1/D19-1259} {{P}ub{M}ed{QA}: A
  dataset for biomedical research question answering}.
\newblock In \emph{Proceedings of the 2019 Conference on Empirical Methods in
  Natural Language Processing and the 9th International Joint Conference on
  Natural Language Processing (EMNLP-IJCNLP)}, pages 2567--2577, Hong Kong,
  China. Association for Computational Linguistics.

\bibitem[{Junczys-Dowmunt et~al.(2018)Junczys-Dowmunt, Grundkiewicz, Dwojak,
  Hoang, Heafield, Neckermann, Seide, Germann, Aji, Bogoychev, Martins, and
  Birch}]{junczys-dowmunt-etal-2018-marian}
Marcin Junczys-Dowmunt, Roman Grundkiewicz, Tomasz Dwojak, Hieu Hoang, Kenneth
  Heafield, Tom Neckermann, Frank Seide, Ulrich Germann, Alham~Fikri Aji,
  Nikolay Bogoychev, Andr{\'e} F.~T. Martins, and Alexandra Birch. 2018.
\newblock \href {https://doi.org/10.18653/v1/P18-4020} {{M}arian: Fast neural
  machine translation in {C}++}.
\newblock In \emph{Proceedings of {ACL} 2018, System Demonstrations}, pages
  116--121, Melbourne, Australia. Association for Computational Linguistics.

\bibitem[{Khashabi et~al.(2018)Khashabi, Chaturvedi, Roth, Upadhyay, and
  Roth}]{khashabi-etal-2018-looking}
Daniel Khashabi, Snigdha Chaturvedi, Michael Roth, Shyam Upadhyay, and Dan
  Roth. 2018.
\newblock \href {https://doi.org/10.18653/v1/N18-1023} {Looking beyond the
  surface: A challenge set for reading comprehension over multiple sentences}.
\newblock In \emph{Proceedings of the 2018 Conference of the North {A}merican
  Chapter of the Association for Computational Linguistics: Human Language
  Technologies, Volume 1 (Long Papers)}, pages 252--262, New Orleans,
  Louisiana. Association for Computational Linguistics.

\bibitem[{Khashabi et~al.(2022)Khashabi, Kordi, and
  Hajishirzi}]{khashabi2022unifiedqav2}
Daniel Khashabi, Yeganeh Kordi, and Hannaneh Hajishirzi. 2022.
\newblock \href {http://arxiv.org/abs/2202.12359} {Unifiedqa-v2: Stronger
  generalization via broader cross-format training}.

\bibitem[{Khot et~al.(2020)Khot, Clark, Guerquin, Jansen, and
  Sabharwal}]{Khot2020QASCAD}
Tushar Khot, Peter Clark, Michal Guerquin, Peter~Alexander Jansen, and Ashish
  Sabharwal. 2020.
\newblock Qasc: A dataset for question answering via sentence composition.
\newblock \emph{ArXiv}, abs/1910.11473.

\bibitem[{Ko{\v{c}}isk{\'y} et~al.(2018)Ko{\v{c}}isk{\'y}, Schwarz, Blunsom,
  Dyer, Hermann, Melis, and Grefenstette}]{kocisky-etal-2018-narrativeqa}
Tom{\'a}{\v{s}} Ko{\v{c}}isk{\'y}, Jonathan Schwarz, Phil Blunsom, Chris Dyer,
  Karl~Moritz Hermann, G{\'a}bor Melis, and Edward Grefenstette. 2018.
\newblock \href {https://doi.org/10.1162/tacl_a_00023} {The {N}arrative{QA}
  reading comprehension challenge}.
\newblock \emph{Transactions of the Association for Computational Linguistics},
  6:317--328.

\bibitem[{Kwiatkowski et~al.(2019)Kwiatkowski, Palomaki, Redfield, Collins,
  Parikh, Alberti, Epstein, Polosukhin, Devlin, Lee, Toutanova, Jones, Kelcey,
  Chang, Dai, Uszkoreit, Le, and Petrov}]{kwiatkowski-etal-2019-natural}
Tom Kwiatkowski, Jennimaria Palomaki, Olivia Redfield, Michael Collins, Ankur
  Parikh, Chris Alberti, Danielle Epstein, Illia Polosukhin, Jacob Devlin,
  Kenton Lee, Kristina Toutanova, Llion Jones, Matthew Kelcey, Ming-Wei Chang,
  Andrew~M. Dai, Jakob Uszkoreit, Quoc Le, and Slav Petrov. 2019.
\newblock \href {https://doi.org/10.1162/tacl_a_00276} {Natural questions: A
  benchmark for question answering research}.
\newblock \emph{Transactions of the Association for Computational Linguistics},
  7:452--466.

\bibitem[{Laban et~al.(2022)Laban, Wu, Murakhovs{'}ka, Liu, and
  Xiong}]{laban-etal-2022-quiz}
Philippe Laban, Chien-Sheng Wu, Lidiya Murakhovs{'}ka, Wenhao Liu, and Caiming
  Xiong. 2022.
\newblock \href {https://doi.org/10.18653/v1/2022.findings-naacl.9} {Quiz
  design task: Helping teachers create quizzes with automated question
  generation}.
\newblock In \emph{Findings of the Association for Computational Linguistics:
  NAACL 2022}, pages 102--111, Seattle, United States. Association for
  Computational Linguistics.

\bibitem[{Lai et~al.(2017)Lai, Xie, Liu, Yang, and Hovy}]{lai-etal-2017-race}
Guokun Lai, Qizhe Xie, Hanxiao Liu, Yiming Yang, and Eduard Hovy. 2017.
\newblock \href {https://doi.org/10.18653/v1/D17-1082} {{RACE}: Large-scale
  {R}e{A}ding comprehension dataset from examinations}.
\newblock In \emph{Proceedings of the 2017 Conference on Empirical Methods in
  Natural Language Processing}, pages 785--794, Copenhagen, Denmark.
  Association for Computational Linguistics.

\bibitem[{Lee et~al.(2021)Lee, Scialom, Yoon, Dernoncourt, and
  Jung}]{lee2021qace}
Hwanhee Lee, Thomas Scialom, Seunghyun Yoon, Franck Dernoncourt, and Kyomin
  Jung. 2021.
\newblock Qace: Asking questions to evaluate an image caption.
\newblock \emph{arXiv preprint arXiv:2108.12560}.

\bibitem[{Lewis et~al.(2020)Lewis, Liu, Goyal, Ghazvininejad, Mohamed, Levy,
  Stoyanov, and Zettlemoyer}]{lewis-etal-2020-bart}
Mike Lewis, Yinhan Liu, Naman Goyal, Marjan Ghazvininejad, Abdelrahman Mohamed,
  Omer Levy, Veselin Stoyanov, and Luke Zettlemoyer. 2020.
\newblock \href {https://doi.org/10.18653/v1/2020.acl-main.703} {{BART}:
  Denoising sequence-to-sequence pre-training for natural language generation,
  translation, and comprehension}.
\newblock In \emph{Proceedings of the 58th Annual Meeting of the Association
  for Computational Linguistics}, pages 7871--7880, Online. Association for
  Computational Linguistics.

\bibitem[{Liang et~al.(2019)Liang, Li, and Yin}]{pmlr-v101-liang19a}
Yichan Liang, Jianheng Li, and Jian Yin. 2019.
\newblock \href {https://proceedings.mlr.press/v101/liang19a.html} {A new
  multi-choice reading comprehension dataset for curriculum learning}.
\newblock In \emph{Proceedings of The Eleventh Asian Conference on Machine
  Learning}, volume 101 of \emph{Proceedings of Machine Learning Research},
  pages 742--757. PMLR.

\bibitem[{Lin(2004)}]{lin-2004-rouge}
Chin-Yew Lin. 2004.
\newblock \href {https://aclanthology.org/W04-1013} {{ROUGE}: A package for
  automatic evaluation of summaries}.
\newblock In \emph{Text Summarization Branches Out}, pages 74--81, Barcelona,
  Spain. Association for Computational Linguistics.

\bibitem[{Lin et~al.(2019)Lin, Tafjord, Clark, and
  Gardner}]{Lin2019ReasoningOP}
Kevin Lin, Oyvind Tafjord, Peter Clark, and Matt Gardner. 2019.
\newblock Reasoning over paragraph effects in situations.
\newblock \emph{ArXiv}, abs/1908.05852.

\bibitem[{Loshchilov and Hutter(2019)}]{loshchilov2018decoupled}
Ilya Loshchilov and Frank Hutter. 2019.
\newblock \href {https://openreview.net/forum?id=Bkg6RiCqY7} {Decoupled weight
  decay regularization}.
\newblock In \emph{International Conference on Learning Representations}.

\bibitem[{Mihaylov et~al.(2018)Mihaylov, Clark, Khot, and
  Sabharwal}]{mihaylov-etal-2018-suit}
Todor Mihaylov, Peter Clark, Tushar Khot, and Ashish Sabharwal. 2018.
\newblock \href {https://doi.org/10.18653/v1/D18-1260} {Can a suit of armor
  conduct electricity? a new dataset for open book question answering}.
\newblock In \emph{Proceedings of the 2018 Conference on Empirical Methods in
  Natural Language Processing}, pages 2381--2391, Brussels, Belgium.
  Association for Computational Linguistics.

\bibitem[{Miller(1995)}]{10.1145/219717.219748}
George~A. Miller. 1995.
\newblock \href {https://doi.org/10.1145/219717.219748} {Wordnet: A lexical
  database for english}.
\newblock \emph{Commun. ACM}, 38(11):39–41.

\bibitem[{Mohammadshahi and
  Henderson(2020)}]{mohammadshahi-henderson-2020-graph}
Alireza Mohammadshahi and James Henderson. 2020.
\newblock \href {https://doi.org/10.18653/v1/2020.findings-emnlp.294}
  {Graph-to-graph transformer for transition-based dependency parsing}.
\newblock In \emph{Findings of the Association for Computational Linguistics:
  EMNLP 2020}, pages 3278--3289, Online. Association for Computational
  Linguistics.

\bibitem[{Mohammadshahi and
  Henderson(2021{\natexlab{a}})}]{10.1162/tacl_a_00358}
Alireza Mohammadshahi and James Henderson. 2021{\natexlab{a}}.
\newblock \href {https://doi.org/10.1162/tacl_a_00358} {{Recursive
  Non-Autoregressive Graph-to-Graph Transformer for Dependency Parsing with
  Iterative Refinement}}.
\newblock \emph{Transactions of the Association for Computational Linguistics},
  9:120--138.

\bibitem[{Mohammadshahi and
  Henderson(2021{\natexlab{b}})}]{mohammadshahi2021syntaxaware}
Alireza Mohammadshahi and James Henderson. 2021{\natexlab{b}}.
\newblock \href {http://arxiv.org/abs/2104.07704} {Syntax-aware graph-to-graph
  transformer for semantic role labelling}.

\bibitem[{Mohammadshahi et~al.(2019)Mohammadshahi, Lebret, and
  Aberer}]{mohammadshahi-etal-2019-aligning}
Alireza Mohammadshahi, R{\'e}mi Lebret, and Karl Aberer. 2019.
\newblock \href {https://doi.org/10.18653/v1/D19-6402} {Aligning multilingual
  word embeddings for cross-modal retrieval task}.
\newblock In \emph{Proceedings of the Beyond Vision and LANguage: inTEgrating
  Real-world kNowledge (LANTERN)}, pages 11--17, Hong Kong, China. Association
  for Computational Linguistics.

\bibitem[{Mohammadshahi et~al.(2022{\natexlab{a}})Mohammadshahi, Nikoulina,
  Berard, Brun, Henderson, and Besacier}]{mohammadshahi2022small100}
Alireza Mohammadshahi, Vassilina Nikoulina, Alexandre Berard, Caroline Brun,
  James Henderson, and Laurent Besacier. 2022{\natexlab{a}}.
\newblock \href {https://aclanthology.org/2022.emnlp-main.571} {{SM}a{LL}-100:
  Introducing shallow multilingual machine translation model for low-resource
  languages}.
\newblock In \emph{Proceedings of the 2022 Conference on Empirical Methods in
  Natural Language Processing}, pages 8348--8359, Abu Dhabi, United Arab
  Emirates. Association for Computational Linguistics.

\bibitem[{Mohammadshahi et~al.(2022{\natexlab{b}})Mohammadshahi, Nikoulina,
  Berard, Brun, Henderson, and Besacier}]{mohammadshahi2022compressed}
Alireza Mohammadshahi, Vassilina Nikoulina, Alexandre Berard, Caroline Brun,
  James Henderson, and Laurent Besacier. 2022{\natexlab{b}}.
\newblock \href {https://aclanthology.org/2022.findings-emnlp.317} {What do
  compressed multilingual machine translation models forget?}
\newblock In \emph{Findings of the Association for Computational Linguistics:
  EMNLP 2022}, pages 4308--4329, Abu Dhabi, United Arab Emirates. Association
  for Computational Linguistics.

\bibitem[{Murakhovs{'}ka et~al.(2022)Murakhovs{'}ka, Wu, Laban, Niu, Liu, and
  Xiong}]{murakhovska-etal-2022-mixqg}
Lidiya Murakhovs{'}ka, Chien-Sheng Wu, Philippe Laban, Tong Niu, Wenhao Liu,
  and Caiming Xiong. 2022.
\newblock \href {https://doi.org/10.18653/v1/2022.findings-naacl.111}
  {{M}ix{QG}: Neural question generation with mixed answer types}.
\newblock In \emph{Findings of the Association for Computational Linguistics:
  NAACL 2022}, pages 1486--1497, Seattle, United States. Association for
  Computational Linguistics.

\bibitem[{Nema and Khapra(2018)}]{nema-khapra-2018-towards}
Preksha Nema and Mitesh~M. Khapra. 2018.
\newblock \href {https://doi.org/10.18653/v1/D18-1429} {Towards a better metric
  for evaluating question generation systems}.
\newblock In \emph{Proceedings of the 2018 Conference on Empirical Methods in
  Natural Language Processing}, pages 3950--3959, Brussels, Belgium.
  Association for Computational Linguistics.

\bibitem[{{Ostermann} et~al.(2018){Ostermann}, {Modi}, {Roth}, {Thater}, and
  {Pinkal}}]{2018arXiv180305223O}
Simon {Ostermann}, Ashutosh {Modi}, Michael {Roth}, Stefan {Thater}, and
  Manfred {Pinkal}. 2018.
\newblock \href {http://arxiv.org/abs/1803.05223} {{MCScript: A Novel Dataset
  for Assessing Machine Comprehension Using Script Knowledge}}.
\newblock \emph{arXiv e-prints}, page arXiv:1803.05223.

\bibitem[{Ostermann et~al.(2019)Ostermann, Roth, and
  Pinkal}]{ostermann-etal-2019-mcscript2}
Simon Ostermann, Michael Roth, and Manfred Pinkal. 2019.
\newblock \href {https://doi.org/10.18653/v1/S19-1012} {{MCS}cript2.0: A
  machine comprehension corpus focused on script events and participants}.
\newblock In \emph{Proceedings of the Eighth Joint Conference on Lexical and
  Computational Semantics (*{SEM} 2019)}, pages 103--117, Minneapolis,
  Minnesota. Association for Computational Linguistics.

\bibitem[{Pan et~al.(2020)Pan, Xie, Feng, Chua, and
  Kan}]{pan-etal-2020-semantic}
Liangming Pan, Yuxi Xie, Yansong Feng, Tat-Seng Chua, and Min-Yen Kan. 2020.
\newblock \href {https://doi.org/10.18653/v1/2020.acl-main.135} {Semantic
  graphs for generating deep questions}.
\newblock In \emph{Proceedings of the 58th Annual Meeting of the Association
  for Computational Linguistics}, pages 1463--1475, Online. Association for
  Computational Linguistics.

\bibitem[{Papineni et~al.(2002)Papineni, Roukos, Ward, and
  Zhu}]{papineni-etal-2002-bleu}
Kishore Papineni, Salim Roukos, Todd Ward, and Wei-Jing Zhu. 2002.
\newblock \href {https://doi.org/10.3115/1073083.1073135} {{B}leu: a method for
  automatic evaluation of machine translation}.
\newblock In \emph{Proceedings of the 40th Annual Meeting of the Association
  for Computational Linguistics}, pages 311--318, Philadelphia, Pennsylvania,
  USA. Association for Computational Linguistics.

\bibitem[{Peyrard et~al.(2017)Peyrard, Botschen, and
  Gurevych}]{peyrard-etal-2017-learning}
Maxime Peyrard, Teresa Botschen, and Iryna Gurevych. 2017.
\newblock \href {https://doi.org/10.18653/v1/W17-4510} {Learning to score
  system summaries for better content selection evaluation.}
\newblock In \emph{Proceedings of the Workshop on New Frontiers in
  Summarization}, pages 74--84, Copenhagen, Denmark. Association for
  Computational Linguistics.

\bibitem[{Popovi{\'c}(2015)}]{popovic-2015-chrf}
Maja Popovi{\'c}. 2015.
\newblock \href {https://doi.org/10.18653/v1/W15-3049} {chr{F}: character
  n-gram {F}-score for automatic {MT} evaluation}.
\newblock In \emph{Proceedings of the Tenth Workshop on Statistical Machine
  Translation}, pages 392--395, Lisbon, Portugal. Association for Computational
  Linguistics.

\bibitem[{Puri et~al.(2020)Puri, Spring, Shoeybi, Patwary, and
  Catanzaro}]{puri-etal-2020-training}
Raul Puri, Ryan Spring, Mohammad Shoeybi, Mostofa Patwary, and Bryan Catanzaro.
  2020.
\newblock \href {https://doi.org/10.18653/v1/2020.emnlp-main.468} {Training
  question answering models from synthetic data}.
\newblock In \emph{Proceedings of the 2020 Conference on Empirical Methods in
  Natural Language Processing (EMNLP)}, pages 5811--5826, Online. Association
  for Computational Linguistics.

\bibitem[{Qi et~al.(2020{\natexlab{a}})Qi, Zhang, Zhang, Bolton, and
  Manning}]{qi2020stanza}
Peng Qi, Yuhao Zhang, Yuhui Zhang, Jason Bolton, and Christopher~D. Manning.
  2020{\natexlab{a}}.
\newblock \href {https://nlp.stanford.edu/pubs/qi2020stanza.pdf} {Stanza: A
  {Python} natural language processing toolkit for many human languages}.
\newblock In \emph{Proceedings of the 58th Annual Meeting of the Association
  for Computational Linguistics: System Demonstrations}.

\bibitem[{Qi et~al.(2020{\natexlab{b}})Qi, Zhang, Zhang, Bolton, and
  Manning}]{qi-etal-2020-stanza}
Peng Qi, Yuhao Zhang, Yuhui Zhang, Jason Bolton, and Christopher~D. Manning.
  2020{\natexlab{b}}.
\newblock \href {https://doi.org/10.18653/v1/2020.acl-demos.14} {{S}tanza: A
  python natural language processing toolkit for many human languages}.
\newblock In \emph{Proceedings of the 58th Annual Meeting of the Association
  for Computational Linguistics: System Demonstrations}, pages 101--108,
  Online. Association for Computational Linguistics.

\bibitem[{Radford et~al.(2019)Radford, Wu, Child, Luan, Amodei, and
  Sutskever}]{radford2019language}
Alec Radford, Jeff Wu, Rewon Child, David Luan, Dario Amodei, and Ilya
  Sutskever. 2019.
\newblock Language models are unsupervised multitask learners.
\newblock In \emph{OpenAI blog}.

\bibitem[{Raffel et~al.(2020)Raffel, Shazeer, Roberts, Lee, Narang, Matena,
  Zhou, Li, and Liu}]{2020t5}
Colin Raffel, Noam Shazeer, Adam Roberts, Katherine Lee, Sharan Narang, Michael
  Matena, Yanqi Zhou, Wei Li, and Peter~J. Liu. 2020.
\newblock \href {http://jmlr.org/papers/v21/20-074.html} {Exploring the limits
  of transfer learning with a unified text-to-text transformer}.
\newblock \emph{Journal of Machine Learning Research}, 21(140):1--67.

\bibitem[{Rajpurkar et~al.(2018)Rajpurkar, Jia, and
  Liang}]{rajpurkar-etal-2018-know}
Pranav Rajpurkar, Robin Jia, and Percy Liang. 2018.
\newblock \href {https://doi.org/10.18653/v1/P18-2124} {Know what you don{'}t
  know: Unanswerable questions for {SQ}u{AD}}.
\newblock In \emph{Proceedings of the 56th Annual Meeting of the Association
  for Computational Linguistics (Volume 2: Short Papers)}, pages 784--789,
  Melbourne, Australia. Association for Computational Linguistics.

\bibitem[{Rajpurkar et~al.(2016)Rajpurkar, Zhang, Lopyrev, and
  Liang}]{rajpurkar-etal-2016-squad}
Pranav Rajpurkar, Jian Zhang, Konstantin Lopyrev, and Percy Liang. 2016.
\newblock \href {https://doi.org/10.18653/v1/D16-1264} {{SQ}u{AD}: 100,000+
  questions for machine comprehension of text}.
\newblock In \emph{Proceedings of the 2016 Conference on Empirical Methods in
  Natural Language Processing}, pages 2383--2392, Austin, Texas. Association
  for Computational Linguistics.

\bibitem[{Rebuffel et~al.(2021)Rebuffel, Scialom, Soulier, Piwowarski,
  Lamprier, Staiano, Scoutheeten, and Gallinari}]{rebuffel-etal-2021-data}
Clement Rebuffel, Thomas Scialom, Laure Soulier, Benjamin Piwowarski, Sylvain
  Lamprier, Jacopo Staiano, Geoffrey Scoutheeten, and Patrick Gallinari. 2021.
\newblock \href {https://doi.org/10.18653/v1/2021.emnlp-main.633}
  {Data-{Q}uest{E}val: A referenceless metric for data-to-text semantic
  evaluation}.
\newblock In \emph{Proceedings of the 2021 Conference on Empirical Methods in
  Natural Language Processing}, pages 8029--8036, Online and Punta Cana,
  Dominican Republic. Association for Computational Linguistics.

\bibitem[{Rei et~al.(2020)Rei, Stewart, Farinha, and
  Lavie}]{rei-etal-2020-comet}
Ricardo Rei, Craig Stewart, Ana~C Farinha, and Alon Lavie. 2020.
\newblock \href {https://doi.org/10.18653/v1/2020.emnlp-main.213} {{COMET}: A
  neural framework for {MT} evaluation}.
\newblock In \emph{Proceedings of the 2020 Conference on Empirical Methods in
  Natural Language Processing (EMNLP)}, pages 2685--2702, Online. Association
  for Computational Linguistics.

\bibitem[{Richardson et~al.(2013)Richardson, Burges, and
  Renshaw}]{richardson-etal-2013-mctest}
Matthew Richardson, Christopher~J.C. Burges, and Erin Renshaw. 2013.
\newblock \href {https://aclanthology.org/D13-1020} {{MCT}est: A challenge
  dataset for the open-domain machine comprehension of text}.
\newblock In \emph{Proceedings of the 2013 Conference on Empirical Methods in
  Natural Language Processing}, pages 193--203, Seattle, Washington, USA.
  Association for Computational Linguistics.

\bibitem[{Rogers et~al.(2020)Rogers, Kovaleva, Downey, and
  Rumshisky}]{Rogers_Kovaleva_Downey_Rumshisky_2020}
Anna Rogers, Olga Kovaleva, Matthew Downey, and Anna Rumshisky. 2020.
\newblock \href {https://doi.org/10.1609/aaai.v34i05.6398} {Getting closer to
  ai complete question answering: A set of prerequisite real tasks}.
\newblock \emph{Proceedings of the AAAI Conference on Artificial Intelligence},
  34(05):8722--8731.

\bibitem[{Rus et~al.(2010)Rus, Wyse, Piwek, Lintean, Stoyanchev, and
  Moldovan}]{rus-etal-2010-first}
Vasile Rus, Brendan Wyse, Paul Piwek, Mihai Lintean, Svetlana Stoyanchev, and
  Christian Moldovan. 2010.
\newblock \href {https://aclanthology.org/W10-4234} {The first question
  generation shared task evaluation challenge}.
\newblock In \emph{Proceedings of the 6th International Natural Language
  Generation Conference}. Association for Computational Linguistics.

\bibitem[{Sakaguchi et~al.(2021)Sakaguchi, Bras, Bhagavatula, and
  Choi}]{10.1145/3474381}
Keisuke Sakaguchi, Ronan~Le Bras, Chandra Bhagavatula, and Yejin Choi. 2021.
\newblock \href {https://doi.org/10.1145/3474381} {Winogrande: An adversarial
  winograd schema challenge at scale}.
\newblock \emph{Commun. ACM}, 64(9):99–106.

\bibitem[{Sap et~al.(2019)Sap, Rashkin, Chen, Le~Bras, and
  Choi}]{sap-etal-2019-social}
Maarten Sap, Hannah Rashkin, Derek Chen, Ronan Le~Bras, and Yejin Choi. 2019.
\newblock \href {https://doi.org/10.18653/v1/D19-1454} {Social {IQ}a:
  Commonsense reasoning about social interactions}.
\newblock In \emph{Proceedings of the 2019 Conference on Empirical Methods in
  Natural Language Processing and the 9th International Joint Conference on
  Natural Language Processing (EMNLP-IJCNLP)}, pages 4463--4473, Hong Kong,
  China. Association for Computational Linguistics.

\bibitem[{Scialom et~al.(2021)Scialom, Dray, Gallinari, Lamprier, Piwowarski,
  Staiano, and Wang}]{scialom2021questeval}
Thomas Scialom, Paul-Alexis Dray, Patrick Gallinari, Sylvain Lamprier, Benjamin
  Piwowarski, Jacopo Staiano, and Alex Wang. 2021.
\newblock Questeval: Summarization asks for fact-based evaluation.
\newblock \emph{arXiv preprint arXiv:2103.12693}.

\bibitem[{Scialom et~al.(2019{\natexlab{a}})Scialom, Lamprier, Piwowarski, and
  Staiano}]{scialom2019answers}
Thomas Scialom, Sylvain Lamprier, Benjamin Piwowarski, and Jacopo Staiano.
  2019{\natexlab{a}}.
\newblock Answers unite! unsupervised metrics for reinforced summarization
  models.
\newblock \emph{arXiv preprint arXiv:1909.01610}.

\bibitem[{Scialom et~al.(2019{\natexlab{b}})Scialom, Piwowarski, and
  Staiano}]{scialom2019self}
Thomas Scialom, Benjamin Piwowarski, and Jacopo Staiano. 2019{\natexlab{b}}.
\newblock Self-attention architectures for answer-agnostic neural question
  generation.
\newblock In \emph{Proceedings of the 57th annual meeting of the Association
  for Computational Linguistics}, pages 6027--6032.

\bibitem[{Sellam et~al.(2020)Sellam, Das, and Parikh}]{sellam-etal-2020-bleurt}
Thibault Sellam, Dipanjan Das, and Ankur Parikh. 2020.
\newblock \href {https://doi.org/10.18653/v1/2020.acl-main.704} {{BLEURT}:
  Learning robust metrics for text generation}.
\newblock In \emph{Proceedings of the 58th Annual Meeting of the Association
  for Computational Linguistics}, pages 7881--7892, Online. Association for
  Computational Linguistics.

\bibitem[{Stanojevi{\'c} and Sima{'}an(2014)}]{stanojevic-simaan-2014-beer}
Milo{\v{s}} Stanojevi{\'c} and Khalil Sima{'}an. 2014.
\newblock \href {https://doi.org/10.3115/v1/W14-3354} {{BEER}: {BE}tter
  evaluation as ranking}.
\newblock In \emph{Proceedings of the Ninth Workshop on Statistical Machine
  Translation}, pages 414--419, Baltimore, Maryland, USA. Association for
  Computational Linguistics.

\bibitem[{Sun et~al.(2019)Sun, Yu, Chen, Yu, Choi, and Cardie}]{dream2019}
Kai Sun, Dian Yu, Jianshu Chen, Dong Yu, Yejin Choi, and Claire Cardie. 2019.
\newblock \href {https://doi.org/10.1162/tacl_a_00264} {Dream: A challenge data
  set and models for dialogue-based reading comprehension}.
\newblock \emph{Transactions of the Association for Computational Linguistics},
  7:217–231.

\bibitem[{Talmor et~al.(2019)Talmor, Herzig, Lourie, and
  Berant}]{talmor-etal-2019-commonsenseqa}
Alon Talmor, Jonathan Herzig, Nicholas Lourie, and Jonathan Berant. 2019.
\newblock \href {https://doi.org/10.18653/v1/N19-1421} {{C}ommonsense{QA}: A
  question answering challenge targeting commonsense knowledge}.
\newblock In \emph{Proceedings of the 2019 Conference of the North {A}merican
  Chapter of the Association for Computational Linguistics: Human Language
  Technologies, Volume 1 (Long and Short Papers)}, pages 4149--4158,
  Minneapolis, Minnesota. Association for Computational Linguistics.

\bibitem[{Thompson and Post(2020)}]{thompson-post-2020-automatic}
Brian Thompson and Matt Post. 2020.
\newblock \href {https://doi.org/10.18653/v1/2020.emnlp-main.8} {Automatic
  machine translation evaluation in many languages via zero-shot paraphrasing}.
\newblock In \emph{Proceedings of the 2020 Conference on Empirical Methods in
  Natural Language Processing (EMNLP)}, pages 90--121, Online. Association for
  Computational Linguistics.

\bibitem[{Trischler et~al.(2016)Trischler, Wang, Yuan, Harris, Sordoni,
  Bachman, and Suleman}]{trischler2016newsqa}
Adam Trischler, Tong Wang, Xingdi Yuan, Justin Harris, Alessandro Sordoni,
  Philip Bachman, and Kaheer Suleman. 2016.
\newblock \href {http://arxiv.org/abs/1611.09830} {Newsqa: A machine
  comprehension dataset}.

\bibitem[{Ushio et~al.(2022)Ushio, Alva-Manchego, and
  Camacho-Collados}]{ushio2022generative}
Asahi Ushio, Fernando Alva-Manchego, and Jose Camacho-Collados. 2022.
\newblock \href {http://arxiv.org/abs/2210.03992} {Generative language models
  for paragraph-level question generation}.

\bibitem[{Ushio et~al.(2023{\natexlab{a}})Ushio, Alva-Manchego, and
  Camacho-Collados}]{ushio-etal-2023-an-empirical}
Asahi Ushio, Fernando Alva-Manchego, and Jose Camacho-Collados.
  2023{\natexlab{a}}.
\newblock An empirical comparison of lm-based question and answer generation
  methods.
\newblock In \emph{Proceedings of the 61th Annual Meeting of the Association
  for Computational Linguistics}, Toronto, Canada. Association for
  Computational Linguistics.

\bibitem[{Ushio et~al.(2023{\natexlab{b}})Ushio, Alva-Manchego, and
  Camacho-Collados}]{ushio-etal-2023-a-practical-toolkit}
Asahi Ushio, Fernando Alva-Manchego, and Jose Camacho-Collados.
  2023{\natexlab{b}}.
\newblock A practical toolkit for multilingual question and answer generation,
  acl 2022, system demonstration.
\newblock In \emph{Proceedings of the 61th Annual Meeting of the Association
  for Computational Linguistics: System Demonstrations}, Toronto, Canada.
  Association for Computational Linguistics.

\bibitem[{Vilares and
  G{\'o}mez-Rodr{\'\i}guez(2019)}]{vilares-gomez-rodriguez-2019-head}
David Vilares and Carlos G{\'o}mez-Rodr{\'\i}guez. 2019.
\newblock \href {https://doi.org/10.18653/v1/P19-1092} {{HEAD}-{QA}: A
  healthcare dataset for complex reasoning}.
\newblock In \emph{Proceedings of the 57th Annual Meeting of the Association
  for Computational Linguistics}, pages 960--966, Florence, Italy. Association
  for Computational Linguistics.

\bibitem[{Wang et~al.(2022)Wang, Liu, Tang, and Wu}]{wang2022qrelscore}
Xiaoqiang Wang, Bang Liu, Siliang Tang, and Lingfei Wu. 2022.
\newblock \href {http://arxiv.org/abs/2204.13921} {Qrelscore: Better evaluating
  generated questions with deeper understanding of context-aware relevance}.

\bibitem[{Wu et~al.(2022)Wu, Madotto, Liu, Fung, and
  Xiong}]{wu-etal-2022-qaconv}
Chien-Sheng Wu, Andrea Madotto, Wenhao Liu, Pascale Fung, and Caiming Xiong.
  2022.
\newblock \href {https://doi.org/10.18653/v1/2022.acl-long.370} {{QAC}onv:
  Question answering on informative conversations}.
\newblock In \emph{Proceedings of the 60th Annual Meeting of the Association
  for Computational Linguistics (Volume 1: Long Papers)}, pages 5389--5411,
  Dublin, Ireland. Association for Computational Linguistics.

\bibitem[{Xiong et~al.(2019)Xiong, Wu, Wang, Kulkarni, Yu, Chang, Guo, and
  Wang}]{xiong-etal-2019-tweetqa}
Wenhan Xiong, Jiawei Wu, Hong Wang, Vivek Kulkarni, Mo~Yu, Shiyu Chang,
  Xiaoxiao Guo, and William~Yang Wang. 2019.
\newblock \href {https://doi.org/10.18653/v1/P19-1496} {{TWEETQA}: A social
  media focused question answering dataset}.
\newblock In \emph{Proceedings of the 57th Annual Meeting of the Association
  for Computational Linguistics}, pages 5020--5031, Florence, Italy.
  Association for Computational Linguistics.

\bibitem[{Yu et~al.(2020)Yu, Jiang, Dong, and Feng}]{Yu2020ReClor}
Weihao Yu, Zihang Jiang, Yanfei Dong, and Jiashi Feng. 2020.
\newblock \href {https://openreview.net/forum?id=HJgJtT4tvB} {Reclor: A reading
  comprehension dataset requiring logical reasoning}.
\newblock In \emph{International Conference on Learning Representations}.

\bibitem[{Yuan et~al.(2021)Yuan, Neubig, and Liu}]{bartscore}
Weizhe Yuan, Graham Neubig, and Pengfei Liu. 2021.
\newblock \href
  {https://proceedings.neurips.cc/paper/2021/file/e4d2b6e6fdeca3e60e0f1a62fee3d9dd-Paper.pdf}
  {Bartscore: Evaluating generated text as text generation}.
\newblock In \emph{Advances in Neural Information Processing Systems},
  volume~34, pages 27263--27277. Curran Associates, Inc.

\bibitem[{Zhang et~al.(2018)Zhang, Liu, Liu, Gao, Duh, and
  Durme}]{zhang2018record}
Sheng Zhang, Xiaodong Liu, Jingjing Liu, Jianfeng Gao, Kevin Duh, and
  Benjamin~Van Durme. 2018.
\newblock \href {http://arxiv.org/abs/1810.12885} {Record: Bridging the gap
  between human and machine commonsense reading comprehension}.

\bibitem[{Zhang and Bansal(2019)}]{zhang-bansal-2019-addressing}
Shiyue Zhang and Mohit Bansal. 2019.
\newblock \href {https://doi.org/10.18653/v1/D19-1253} {Addressing semantic
  drift in question generation for semi-supervised question answering}.
\newblock In \emph{Proceedings of the 2019 Conference on Empirical Methods in
  Natural Language Processing and the 9th International Joint Conference on
  Natural Language Processing (EMNLP-IJCNLP)}, pages 2495--2509, Hong Kong,
  China. Association for Computational Linguistics.

\bibitem[{Zhang et~al.(2020)Zhang, Kishore, Wu, Weinberger, and
  Artzi}]{Zhang2020BERTScore}
Tianyi Zhang, Varsha Kishore, Felix Wu, Kilian~Q. Weinberger, and Yoav Artzi.
  2020.
\newblock \href {https://openreview.net/forum?id=SkeHuCVFDr} {Bertscore:
  Evaluating text generation with bert}.
\newblock In \emph{International Conference on Learning Representations}.

\bibitem[{Zhao et~al.(2019)Zhao, Peyrard, Liu, Gao, Meyer, and
  Eger}]{zhao-etal-2019-moverscore}
Wei Zhao, Maxime Peyrard, Fei Liu, Yang Gao, Christian~M. Meyer, and Steffen
  Eger. 2019.
\newblock \href {https://doi.org/10.18653/v1/D19-1053} {{M}over{S}core: Text
  generation evaluating with contextualized embeddings and earth mover
  distance}.
\newblock In \emph{Proceedings of the 2019 Conference on Empirical Methods in
  Natural Language Processing and the 9th International Joint Conference on
  Natural Language Processing (EMNLP-IJCNLP)}, pages 563--578, Hong Kong,
  China. Association for Computational Linguistics.

\bibitem[{Zhou et~al.(2017)Zhou, Yang, Wei, Tan, Bao, and
  Zhou}]{https://doi.org/10.48550/arxiv.1704.01792}
Qingyu Zhou, Nan Yang, Furu Wei, Chuanqi Tan, Hangbo Bao, and Ming Zhou. 2017.
\newblock \href {https://doi.org/10.48550/ARXIV.1704.01792} {Neural question
  generation from text: A preliminary study}.

\end{thebibliography}
